%% file: main.tex
\documentclass{article}
\usepackage{iftex}
\ifPDFTeX
  \pdfoutput=1
\fi
\usepackage{colm2024_conference}
\input{math_commands.tex}

\usepackage{amssymb}
\usepackage{graphicx}
\usepackage{enumitem}
\usepackage{booktabs}
\usepackage{multirow}
\usepackage{array,tikz}
\usepackage{marvosym}
\input{tabular/main_style}

\title{%
\texorpdfstring{%
  \raisebox{-0.15em}{%
    \includegraphics[height=2.5em]{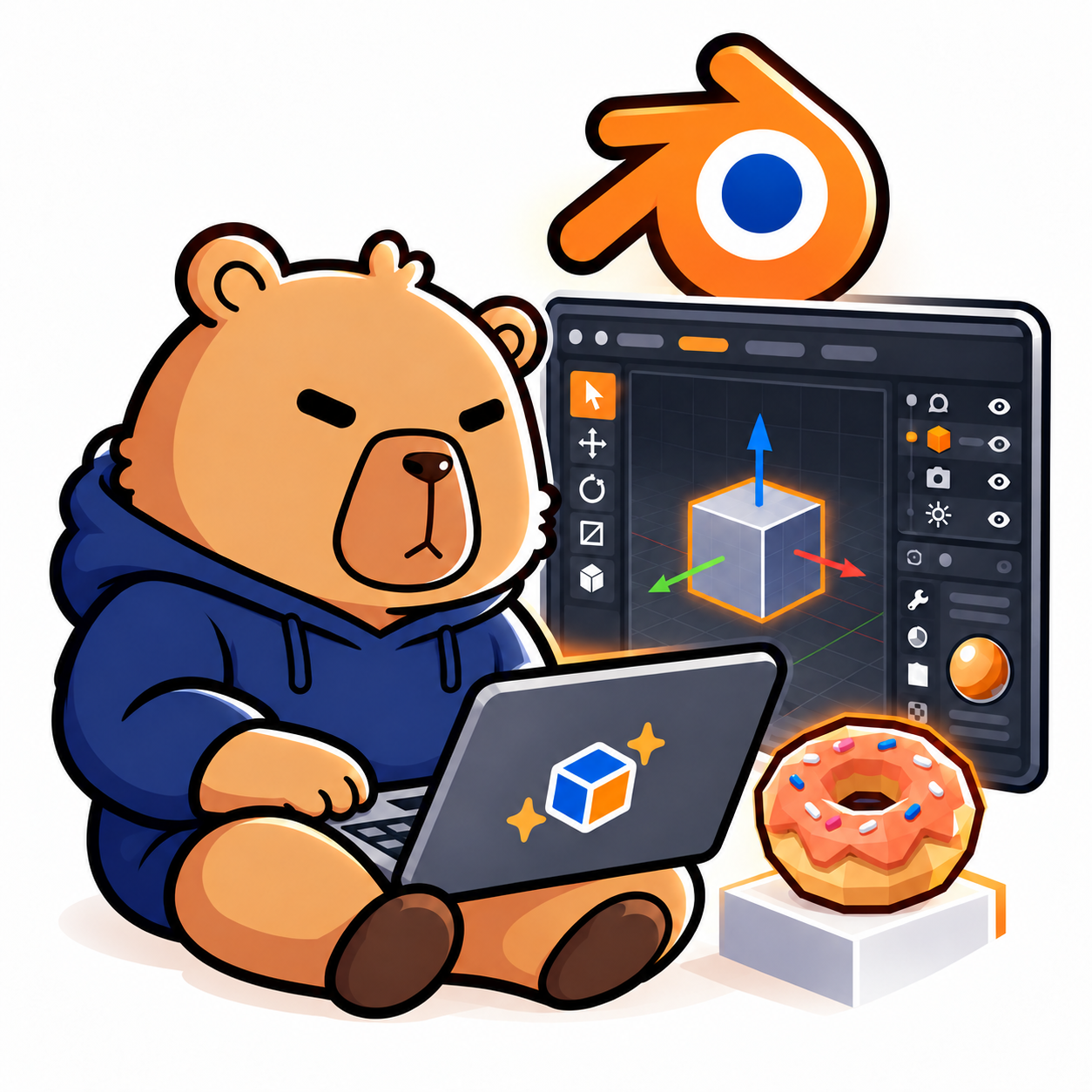}%
  }\hspace{0.3em}%
}{}%
BlenderWorld: Benchmarking Visual Agents for 3D Creation
from Real-World References
}
\input{authors}
\hypersetup{
  pdftitle={BlenderWorld: Benchmarking Visual Agents for 3D Creation from Real-World References},
  pdfauthor={Xinyu Fang, Yuchi Wang, Weihao Yu, Rongyao Fang, Zhaohai Li, Zhixun Li, Tiancheng Gu, Xilin Wei, Yuhang Zang, Hongsheng Li, Dahua Lin, Guofeng Zhang, Shuai Bai, Zhibo Yang, Dayiheng Liu}
}

\newcommand{\benchname}{\textbf{\textsc{BlenderWorld}}}
\definecolor{halinkcol}{HTML}{1B4F9C}

\begin{document}
\maketitle



\begin{abstract}
Recent advances in multimodal large language models (MLLMs) have enabled them to act as autonomous agents, creating and editing 3D models via code or direct software manipulation. In practical workflows, these models must infer object shapes, materials, spatial layouts, and lighting from diverse references while strictly following instructions to reconstruct, modify, or design 3D content. However, existing benchmarks provide limited coverage of such open-ended requests across real-world scenarios. To bridge this gap, we introduce \benchname{}, the first comprehensive benchmark dedicated to evaluating frontier models on realistic 3D creation tasks. \benchname{} comprises 223 tasks spanning architecture, products, natural environments, and scientific visualization, driven by text, image, video, and structured specification inputs. To rigorously assess these complex deliverables, we establish a unified evaluation protocol: we render the submitted Blender projects from multiple viewpoints and evaluate them against 2,651 task-specific binary rubrics alongside holistic visual quality scores. Benchmarking thirteen frontier models reveals that GPT-6-Astra achieves the highest overall fused score of 79.9, followed by Opus-5 (73.0) and GPT-5.6-Sol (67.7). Ultimately, our findings reveal a critical bottleneck: while current models can parse broad instructions, they still profoundly struggle with fine-grained 3D spatial coherence and accurate structural reproduction in Blender, leaving substantial headroom for future 3D-native agentics.
\end{abstract}

\input{section/1_intro}

\input{section/2_related_work}

\input{section/3_benchmark}

\input{section/4_eval}

\input{section/5_experiment}

\input{section/6_conclusion}

\input{section/7_future}

\bibliography{references}
\bibliographystyle{colm2024_conference}

\input{section/99_appendix}

\end{document}

%% file: math_commands.tex
\usepackage{amsmath,amsfonts,bm}

\def\eqref#1{equation~\ref{#1}}

\def\1{\bm{1}}

\DeclareMathAlphabet{\mathsfit}{\encodingdefault}{\sfdefault}{m}{sl}
\SetMathAlphabet{\mathsfit}{bold}{\encodingdefault}{\sfdefault}{bx}{n}



%% file: tabular/main_style.tex
\definecolor{bwscorewash}{HTML}{5A8BFC}
\definecolor{bwtokenwash}{HTML}{E4AD4B}
\definecolor{bwtableink}{HTML}{232326}
\definecolor{bwseparator}{HTML}{CBCCD4}

\newcolumntype{B}{!{\color{bwseparator}\vrule width 0.35pt}}
\newcommand{\bwbarheight}{0.29}
\newcommand{\bwbarfont}{\fontsize{8}{9}\selectfont}

\newcommand{\bwtablebar}[4][0.84]{%
  \begin{tikzpicture}[baseline=0.40ex,x=1cm,y=1cm]
    \fill[#2!10,rounded corners=0.8pt] (0,0) rectangle (#1,\bwbarheight);
    \shade[left color=#2!47,right color=#2!18,rounded corners=0.8pt]
      (0,0) rectangle ({#4*#1},\bwbarheight);
    \node[inner sep=0pt,text=bwtableink] at ({#1/2},{\bwbarheight/2})
      {\bwbarfont #3};
  \end{tikzpicture}%
}
\newcommand{\bwtablebest}[4][0.84]{%
  \begin{tikzpicture}[baseline=0.40ex,x=1cm,y=1cm]
    \fill[#2!10,rounded corners=0.8pt] (0,0) rectangle (#1,\bwbarheight);
    \shade[left color=#2!59,right color=#2!26,rounded corners=0.8pt]
      (0,0) rectangle ({#4*#1},\bwbarheight);
    \draw[#2!64,rounded corners=0.8pt,line width=0.4pt]
      (0,0) rectangle (#1,\bwbarheight);
    \node[inner sep=0pt,text=bwtableink] at ({#1/2},{\bwbarheight/2})
      {\bwbarfont\bfseries #3};
  \end{tikzpicture}%
}

\newcommand{\bwmodel}[2]{%
  \makebox[0.34cm][c]{\raisebox{-0.15\height}{%
    \includegraphics[width=0.26cm,height=0.26cm,keepaspectratio]{tabular/icons/#1.pdf}}}%
  \hspace{0.25em}{\ttfamily\fontsize{7.6}{9}\selectfont #2}%
}
\newsavebox{\bwmaintablebox}

%% file: authors.tex
\author{%
  {\normalsize\bfseries
  \begin{tabular}{@{}c@{}}
    Xinyu Fang\textsuperscript{1,2}\quad
    Yuchi Wang\textsuperscript{2,3}\quad
    Weihao Yu\textsuperscript{2,3}\quad
    Rongyao Fang\textsuperscript{2}\quad
    Zhaohai Li\textsuperscript{2}\\[0.35em]
    Zhixun Li\textsuperscript{2,3}\quad
    Tiancheng Gu\textsuperscript{2}\quad
    Xilin Wei\textsuperscript{2,4}\quad
    Yuhang Zang\textsuperscript{5}\quad
    Hongsheng Li\textsuperscript{3}\\[0.35em]
    Dahua Lin\textsuperscript{3,\textdagger}\quad
    Guofeng Zhang\textsuperscript{1,\textdagger}\quad
    Shuai Bai\textsuperscript{2}\quad
    Zhibo Yang\textsuperscript{2,\textdagger}\quad
    Dayiheng Liu\textsuperscript{2}
  \end{tabular}}\\[0.75em]
  {\small\normalfont
  \begin{tabular}{@{}c@{}}
    \textsuperscript{1}Zhejiang University\quad
    \textsuperscript{2}Alibaba Token Hub, Alibaba Group\quad
    \textsuperscript{3}The Chinese University of Hong Kong\\
    \textsuperscript{4}Fudan University\quad
    \textsuperscript{5}National University of Singapore
    \end{tabular}}\\[0.3em]
    {\small\normalfont
      \texttt{\Letter\,fangxinyu09@zju.edu.cn, zhibo.yzb@alibaba-inc.com}
    }\\[0.3em]
    {\small\normalfont
      \faGlobe\, Website: \href{https://blenderworld.github.io/}{\textcolor{halinkcol}{\texttt{blenderworld.github.io}}}
    }%
  }

%% file: section/1_intro.tex
\section{Introduction}

Visual agents built on multimodal large language models (MLLMs) can interpret visual references, operate design tools, and refine their outputs through visual inspection~\citep{qwen38max,openai2026gpt6}. These agents have gained traction in two-dimensional (2D) design tasks, including webpage, poster, layout, and presentation creation~\citep{si2024design2code,chen2024textlap,li2025sketch2code,zhao2026unipptbench}, and are increasingly being extended to three-dimensional (3D) modeling and scene creation.
3D modeling is a crucial task and provides a practically valuable testbed for visual agents. First, it involves complex, long-horizon workflows in which agents must coordinate extensive tool use across dozens of interaction turns. Second, it supports applications ranging from everyday personal projects to professional scientific visualization, while spanning scales from individual household objects to entire city blocks. Third, mastering this domain requires spatial reasoning, active visual verification, and sustained tool manipulation within open-ended digital environments, all of which are essential for extending the productivity frontier of AI agents.

However, evaluating agents on practical 3D modeling tasks presents three major challenges: environment selection, task coverage, and output assessment. First, evaluation requires an accessible environment that supports 3D creation, editing, and inspection, yet few available environments provide all three capabilities. We therefore turn to Blender, an open-source platform that supports these capabilities through Python scripting and Model Context Protocol (MCP) interfaces and has been used for agent-driven creation and editing~\citep{hu2024scenecraftllmagentsynthesizing,huang2024blenderalchemyediting3dgraphics,noor2026nova3dcodenativegenerationprogrammable}. Second, practical requests vary widely in their reference sources and design goals, ranging from reconstruction based on photographs or illustrations to open-ended design briefs. A realistic benchmark must therefore cover diverse application contexts and user requirements. Existing Blender-based benchmarks typically focus on narrower task classes: 3DCodeBench and 3DHarnessBench emphasize individual objects~\citep{gao20263dcodebenchbenchmarkingagenticprocedural,liu20263dharnessbench}, BlenderBench contains a small set of camera and scene-editing tasks~\citep{yin2026visionasinversegraphicsagentinterleavedmultimodal}, and SceneActBench covers broader scenes but focuses on predefined geometric objectives~\citep{zhao2026sceneactbench}. Finally, realistic requests often lack ground-truth 3D assets, while open-ended briefs may admit multiple valid solutions. Reliable evaluation must therefore assess both visual properties and underlying structures, requiring evidence beyond what a single rendered view can reveal.

\begin{figure}[t]
    \centering
    \includegraphics[width=\linewidth]{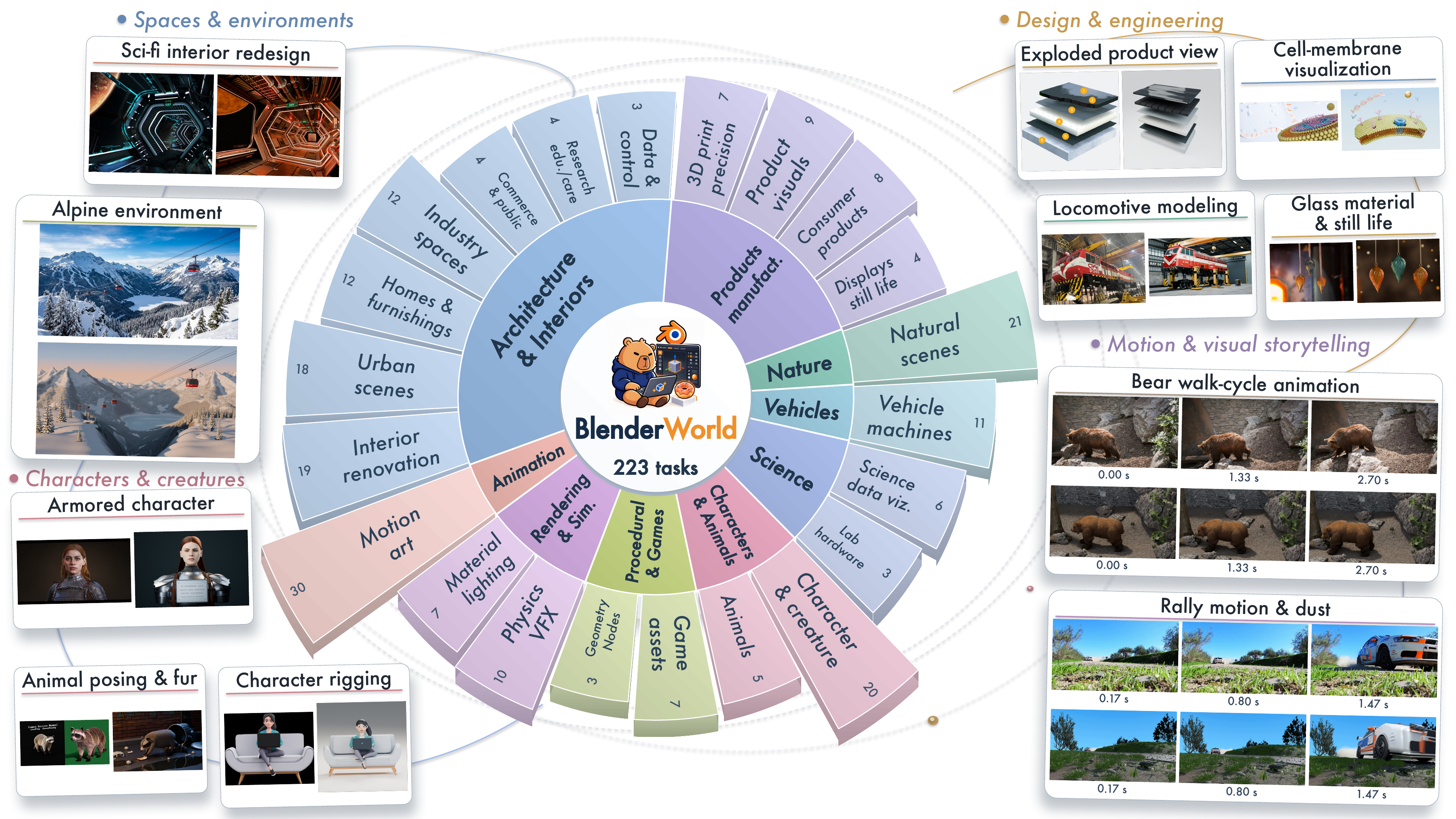}
    \caption{\benchname{} taxonomy and examples. References are shown left/top and GPT-6-Astra results right/bottom.}
    \label{fig:blenderworld_intro}
\end{figure}
\vspace{-5pt}

To address these challenges, we introduce \benchname{}, a benchmark for evaluating visual agents on realistic 3D reconstruction, modification, and design tasks in Blender. \benchname{} contains 223 tasks spanning 16 practical application scenarios, organized into 9 main categories and 22 subcategories (Fig.~\ref{fig:blenderworld_intro}). The tasks represent diverse user needs, driven by text, image, and video references, with 59 tasks additionally including structured specification files. Agents must synthesize these inputs to produce fully editable Blender projects. To rigorously evaluate these diverse deliverables without a single canonical asset, we design an evidence-routed evaluation protocol. An automatic agent selects and renders multiple task-conditioned views, which are then assessed via 2,651 task-specific binary rubrics and continuous visual-quality dimensions. Crucially, we employ a dual-judge system: a VLM-as-Judge for visual and spatial properties, and an Agent-as-Judge for direct programmatic inspection of scene structures that bypass visual shortcuts. Benchmarking 13 frontier models reveals a stark reality: while models can generate visually plausible outputs and plan extensive tool invocations, they profoundly struggle to translate passive visual feedback into structurally sound 3D corrections.

\vspace{-10pt}
\paragraph{Contributions.}
We summarize our main contributions as follows:
\textbf{(1)}~We introduce \textsc{BlenderWorld}, a comprehensive benchmark of 223 realistic Blender tasks spanning 22 subcategories, covering diverse text, image, and video reference sources alongside specifications for practical 3D reconstruction, modification, and design requests.
\textbf{(2)}~We provide a unified evaluation protocol that combines task-conditioned multi-view rendering, task-specific binary rubrics (verified via both visual judges and programmatic scene-state inspection), and a holistic visual quality assessment.
\textbf{(3)}~We evaluate agents built on thirteen frontier models under a shared task and protocol, establishing strong baselines. Our results reveal that while top-performing models like GPT-6-Astra achieve high overall fused scores (79.9), they still fail to fully satisfy all task requirements, reaching a perfect pass rate on only 42.2\% of tasks.
\textbf{(4)}~Through in-depth trajectory and failure analysis, we expose fundamental bottlenecks in current 3D agents: they frequently rely on 2D visual shortcuts (e.g., view-dependent geometry) and profoundly struggle with fine-grained 3D spatial coherence, physical contact constraints, and translating passive visual inspection into effective structural refinements.

%% file: section/2_related_work.tex
\section{Related Work}
\input{tabular/comparison}

\textbf{Productivity-Oriented and Real-World Agent Benchmarks.}
Productivity-oriented benchmarks assess whether agents can complete software workflows and produce verifiable deliverables. SaaS-Bench, WildClawBench, and Claw-Eval examine professional workflows \citep{shi2026saasbench,ding2026wildclawbench,ye2026claweval}, while WebArena, WorkArena, and OSWorld provide browser or desktop tasks \citep{zhou2023webarena,drouin2024workarena,xie2024osworld}. \benchname{} extends this focus on practical deliverables to everyday 3D production briefs, including renovation proposals, product visualization, and classroom teaching. Briefs may include photographs, illustrations, videos, and separate specification files. Agents must interpret these inputs together with the user's instructions and deliver editable Blender projects with the requested media.

\textbf{Procedural 3D and CAD Benchmarks.}
Existing Blender-based benchmarks cover object modeling \citep{gao20263dcodebenchbenchmarkingagenticprocedural,liu20263dharnessbench}, scene editing \citep{gu2025blendergymbenchmarkingfoundationalmodel,yin2026visionasinversegraphicsagentinterleavedmultimodal}, and scene reconstruction or dynamics \citep{yang2026codeasroomgenerating3drooms,zhao2026sceneactbench,liu2026simworldsmultiagentdynamic3d}. CADBench, BenchCAD, and MUSE emphasize geometric correctness, engineering intent, or assembly \citep{doris2026cadbenchmultimodalbenchmarkaiassisted,zhang2026benchcadcomprehensiveindustrystandardbenchmark,dong2026musebenchmarkingmanufacturablefunctional}. Table~\ref{tab:benchmark_comparison} compares task coverage and artifact evaluation across representative 3D benchmarks. \benchname{} combines reconstruction, modification, and reference-based design across 9 task categories and 16 application scenarios. Generation agents iteratively build editable projects, while evaluator agents select task-relevant views and inspect scene structure. Task-specific binary rubrics assess compliance with each brief, while visual quality scores assess the overall rendered result.

%% file: tabular/comparison.tex
\begin{table*}[t]
\centering
\caption{\textbf{Task coverage and artifact evaluation in programmatic 3D benchmarks.}
\benchname{} spans nine task categories, combining
multimodal references with task-specific rubrics,
agent-selected views, and agent inspection of editable projects.}
\label{tab:benchmark_comparison}

\begingroup
\hypersetup{hidelinks}
\providecolor{cdgreen}{HTML}{397D61}
\providecolor{cdink}{HTML}{25342F}
\providecolor{cdmuted}{HTML}{87928B}
\colorlet{bwtableyes}{cdgreen!70!cdink}
\colorlet{bwtableno}{cdmuted!65!cdink}
\colorlet{bwtablecite}{cdink!72}
\newcommand{\bwcheck}{\textcolor{bwtableyes}{\ensuremath{\checkmark}}}
\newcommand{\bwcross}{\textcolor{bwtableno}{\ensuremath{\times}}}
\newcommand{\bwnone}{\textcolor{bwtableno}{---}}
\newcommand{\bwbenchcite}[2]{#1\,{\scriptsize
  \textcolor{bwtablecite}{\citep{#2}}}}
\small
\setlength{\tabcolsep}{3.5pt}
\renewcommand{\arraystretch}{1.1}
\resizebox{\textwidth}{!}{%
\begin{tabular}{@{}lccc@{\hspace{12pt}}ccc@{}}
\toprule
\multirow{3}{*}{\textbf{Benchmark}}
  & \multicolumn{3}{c}{\textbf{Task}}
  & \multicolumn{3}{c}{\textbf{Evaluation}} \\
\cmidrule(lr){2-4}\cmidrule(lr){5-7}
  & \textbf{Task}
  & \textbf{Tasks}
  & \textbf{Input}
  & \textbf{Task-specific}
  & \textbf{Visual}
  & \textbf{3D} \\
  & \textbf{scope}
  & \textbf{num}
  & \textbf{modes}
  & \textbf{rubrics}
  & \textbf{evidence}
  & \textbf{checks} \\
\midrule
\bwbenchcite{3DCodeBench}{gao20263dcodebenchbenchmarkingagenticprocedural} & Objects & 212 & T,I & \bwnone & Fixed & Geom. \\
\bwbenchcite{3DHarnessBench}{liu20263dharnessbench} & Objects & 100 & I,3D & \bwnone & Fixed & Geom. \\
\bwbenchcite{VoxelCodeBench}{zheng2026voxelcodebenchbenchmarking3dworld} & Voxel scenes & 220 & T & \bwnone & MV & \bwnone \\
\bwbenchcite{BlenderGym}{gu2025blendergymbenchmarkingfoundationalmodel} & Scene edits & 245 & T,I & \bwnone & Fixed & Geom. \\
\bwbenchcite{BlenderBench}{yin2026visionasinversegraphicsagentinterleavedmultimodal} & Camera / edits & 27 & T,I & \bwnone & Renders & \bwnone \\
\bwbenchcite{Code-as-Room}{yang2026codeasroomgenerating3drooms} & Rooms & 41 & I & \bwnone & Renders & Layout \\
\bwbenchcite{SceneActBench}{zhao2026sceneactbench} & Scene actions & 520 & I,V & \bwnone & \bwnone & Geom. \\
\bwbenchcite{4DBuildBench}{liu2026simworldsmultiagentdynamic3d} & Dynamics & 50 & T & \bwcheck & Frames & State \\
\bwbenchcite{PhysCodeBench}{xie2026physcodebenchbenchmarkingphysicsawaresymbolic} & Physics & 200 & T & \bwcheck & Video & Physics \\
\bwbenchcite{WorldCoder-Bench (Core)}{lu2026worldcoderbenchbenchmarkingphysicallygrounded} & 3D worlds & 205 & T & \bwcheck & \bwnone & Contracts \\
\midrule
\textbf{BlenderWorld (ours)} & \textbf{9 categories} & \textbf{223} & \textbf{T,I,V} & \textbf{2,651} & \textbf{Agent} & \textbf{Agent} \\
\bottomrule
\end{tabular}%
}

\vspace{3pt}
\begin{minipage}{\textwidth}
\scriptsize
T/I/V/3D~=text/image/video/native 3D reference, visual modes may also include instructions. Fixed~=fixed views, MV~=multiple views, Frames~=sampled video frames,
Geom.~=geometric metrics, State~=scripted scene-state audit.
\end{minipage}
\endgroup
\end{table*}

%% file: section/3_benchmark.tex
\vspace{-5pt}
\section{\benchname}
\vspace{-5pt}
\subsection{Task Definition}
\label{sec:task}

\benchname{} evaluates interactive 3D creation from production briefs.
Each task provides $x=(q,I,F)$: a natural-language request $q$, optional image
or video references $I$, and optional specification files $F$. Starting from
workspace state $s_0$, an agent $\pi$ uses observations $o_t$ to select actions:
\vspace{-5pt}
\begin{equation}
    a_t=\pi(x,o_{\leq t}), \qquad s_{t+1}=\mathcal{E}_{\mathrm{B}}(s_t,a_t),
\end{equation}
where $\mathcal{E}_{\mathrm{B}}$ updates the Blender workspace state $s_t$
after action $a_t$. Through Blender's Python API and MCP tools, actions include scene inspection, Python execution, asset import, and rendering. Rendered views and scene queries guide revisions to geometry, materials, lighting, and animation within the interaction budget.
After $T$ steps, the agent outputs $A_T=(B,R,V,X)$: an editable Blender project $B$, renders $R$, animations/videos $V$, and auxiliary files $X$. Deliverables must support downstream reuse as specified. Our evaluation checks these against the brief and references, accommodating multiple valid 3D implementations.

\begin{figure}[t]
    \centering
    \includegraphics[width=\linewidth]{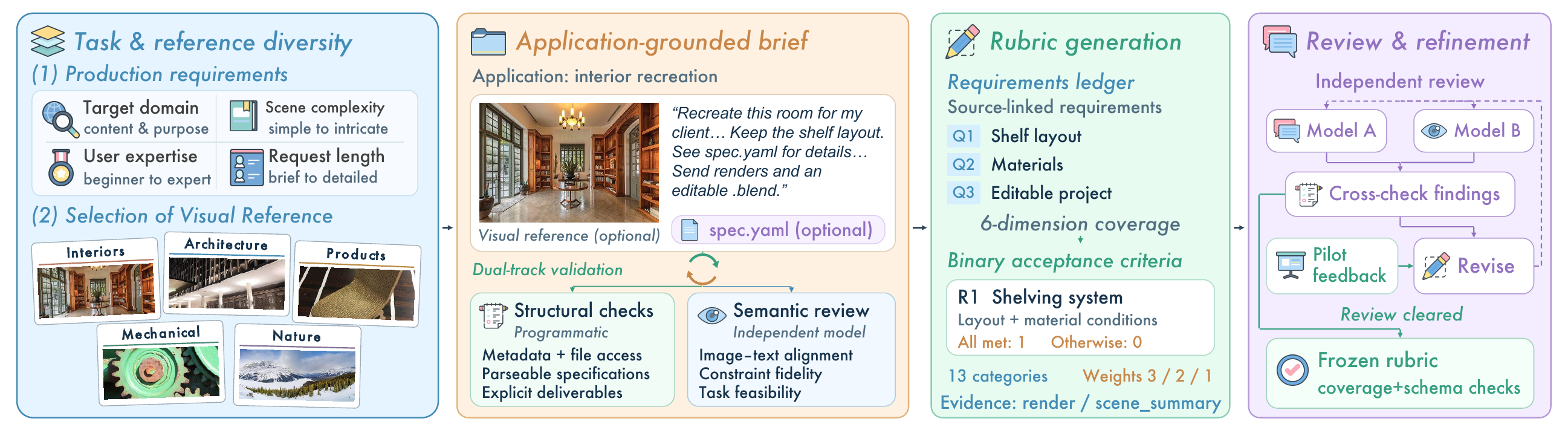}
    \vspace{-1em}
    \caption{\textbf{Construction and rubric derivation pipeline in \benchname{}.} 
Illustrating the end-to-end workflow from diversified task and reference curation, application-grounded brief specification with dual-track validation, to multi-model verified and frozen binary rubrics.}
    \label{fig:construct}
    \vspace{-1em}
\end{figure}

\subsection{Benchmark Construction}
\label{sec:benchmark-construction}

Inspired by practical uses of 3D content in social media, we organize task coverage along two dimensions: category taxonomy and application scenarios. These dimensions guide
task specification and MLLM-assisted brief generation. We then check
the resulting briefs for structural and semantic consistency before
deriving task-specific rubrics (Figure~\ref{fig:construct}).
\vspace{-1em}
\paragraph{Task and reference diversity.}
Task design varies both the production requirements and how references guide the output. \textbf{(1) Production requirements.} We vary the target domain, scene complexity, user expertise, and request length. For example, a teaching task may briefly describe a biological cell with multiple internal structures.
\textbf{(2) Reference use.} We curate diverse reference images spanning interiors, architecture, products, mechanical structures, and natural environments. An MLLM then assesses image quality and structural complexity and we favor images with clearly identifiable structures and detailed information about shape, layout, or materials.
We assign each visual reference one of three roles: \emph{recreation} preserves the main shapes and spatial relationships, \emph{modification} makes requested changes while retaining specified structures, and \emph{reference-based design} draws on the depicted form, style, or layout to develop a new design. The reference image and its assigned role guide both initial task generation and request rewriting.
\vspace{-1em}
\paragraph{Application-grounded specification.}
Given the production target and reference role, we situate each specification in a realistic application scenario. For example, construction reports emphasize structural relationships, whereas product presentations emphasize visibility and appearance. We express each specification as a natural-language request that states the production goal and key requirements. Tasks with detailed specifications may also include \texttt{Markdown}, \texttt{YAML}, or \texttt{JSON} attachments containing dimensions, material lists, camera settings, or asset budgets. The request explicitly references these files, which retain the details required to complete the task.
\vspace{-1em}
\paragraph{Dual-track validation.}
We validate each brief for structural and semantic consistency. \textbf{Structural checks} ensure that the metadata are compatible, images and attachments are accessible, structured files can be parsed, and the brief explicitly requests both rendered outputs and an editable project. For \textbf{semantic review}, a model distinct from the task generator independently examines the initial specification, rewritten request, reference image, and attachments together. It checks image--text alignment, constraint consistency, preservation of key requirements, and task feasibility. Tasks that fail validation are fed back into the generation process for revision.
\vspace{-1em}
\paragraph{Rubric generation.}
A generator extracts requirements from validated briefs, converting them into binary criteria with source links. Each criterion is assigned to one of 13 categories, weighted (3/2/1 for major/secondary/detail), and designated an evidence type: \texttt{render} (for visual properties) or \texttt{scene\_summary} (for project/file inspection). We iteratively refine drafts via pilot evaluations (using GPT-5.6-Sol) to flag ambiguities. Next, at least two independent reviewer models cross-check coverage, source support, and evaluability using concrete counterexamples. An editor resolves flagged issues, with any semantic changes triggering re-review, before freezing the final schema (details in Appendix~\ref{app:rubric_generation}).

\subsection{Dataset Statistics}
\label{sec:taxonomy_statistics}

\begin{figure}[!t]
  \centering
  \setlength{\abovecaptionskip}{10pt}
  \includegraphics[width=\linewidth]{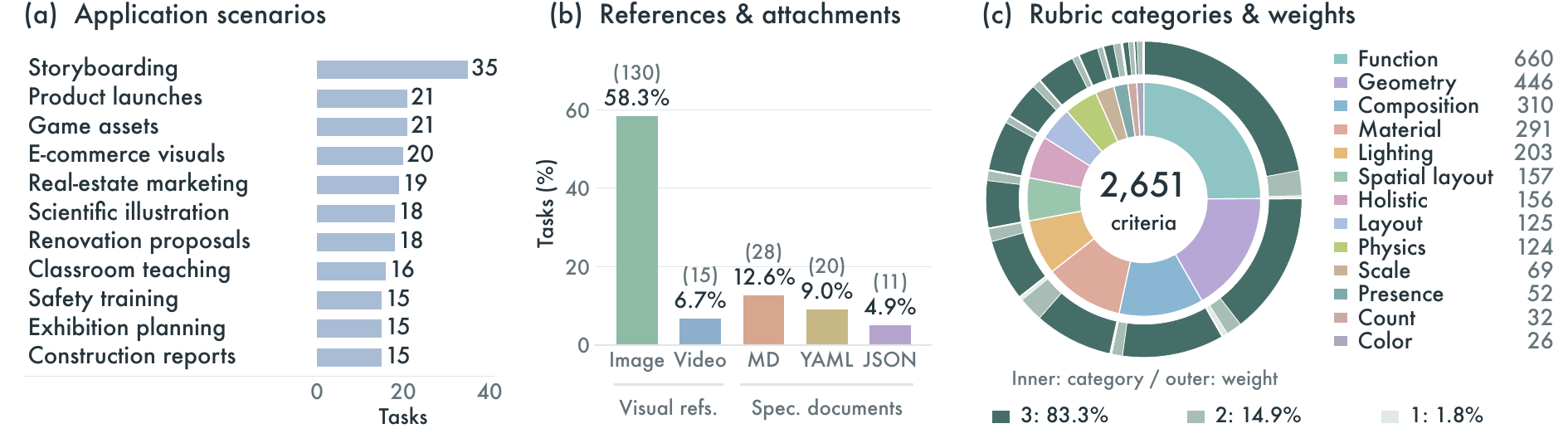}
  \caption{Dataset statistics of \benchname{}. (a) The 11 largest application scenarios (213 of 223 tasks). (b) Percentages of tasks with each reference or specification type; types may co-occur. (c) Criterion counts by rubric category (inner ring) and importance (outer ring).}
  \label{fig:blenderworld_statistics}
  \vspace{-1em}
\end{figure}

\benchname{} contains 223 production briefs across 9 main task
categories and 22 subcategories (Figure~\ref{fig:blenderworld_intro}), covering
architecture and interiors, products and manufacturing, vehicles and machinery,
characters and animals, nature and landscapes, and scientific visualization.
The tasks serve 16 application scenarios, with the largest shown in
Figure~\ref{fig:blenderworld_statistics}a and the full distribution in
Appendix~\ref{app:application_contexts}. Visual references are provided for
145 tasks (65.0\%): 130 include images, and 15 include videos.
Specification files in Markdown, YAML, or JSON accompany 59 tasks (26.5\%),
adding requirements on dimensions, materials, and camera settings
(Figure~\ref{fig:blenderworld_statistics}b). References and specifications can
co-occur, combining visual guidance with explicit production constraints.

Evaluation uses 2,651 binary criteria across 13 rubric categories, with 6--22
criteria per task (mean 11.9).
Function and geometry together account for 41.7\% of criteria. Of all
criteria, 83.3\% have major importance (weight 3). For judging, 41.8\% of
criteria are assigned to rendered views and 58.2\% to scene inspection. Explicit requirements from
specification files account for 500 criteria (18.9\%), linking these
attachments to evaluation.
Table~\ref{tab:benchmark_comparison} compares \benchname{} with existing 3D benchmarks.

%% file: section/4_eval.tex
\section{Evaluation and Metrics}
\label{sec:evaluation}
\subsection{Evaluation Protocol}
\label{sec:protocol}
For each task, an agent uses a harness equipped with Blender MCP to review the task brief and any provided references or attachments,
construct the scene, and save the requested deliverables. We archive the
agent's final workspace, including the editable \texttt{.blend} project,
exported media, and auxiliary files. This artifact package is then processed 
through automatic agent rendering to produce more visual evidence, and finally to get rubric and dimension scores with model judgment.

\paragraph{Automatic Agent Render (AaR).}
Given the task query, rubrics, reference files, and submitted deliverables,
AaR selects eight complementary views to reveal task-relevant details.
This adaptive view selection aims to make the required properties observable
while maintaining a common rendering budget across submissions.
The rendering agent opens the submitted Blender projects and renders each
view at $1024\times1024$ resolution with 64 samples using Cycles.
It may adjust cameras, framing, and rendering settings, but must preserve
the submitted geometry, materials, lighting, animation, and color management.
Invalid views can be re-rendered within two attempts, but defects in the submitted work must
not be repaired. Visual judging proceeds only when newly rendered views are
available; otherwise, the evaluation is marked as failed.
The original exported media remain available as additional evidence
alongside the evaluator-generated views.
\paragraph{Evidence-routed rubric judging.}
Each rubric specifies a binary criterion, an evidence type, and an importance
weight. We route criteria to two complementary judges according to the
evidence needed to verify them. \textbf{Agent-as-Judge} handles \texttt{scene\_summary} criteria, which cover
requirements on internal scene structure and native-file properties that
cannot be reliably verified from rendered views alone. The agent Judger
inspects the deliverables through multiple tools provided by Blender and records
evidence about internal states. Judger also consults visual evidence when necessary and emits a binary decision with
supporting evidence for each assigned criterion. \textbf{VLM-as-Judge} handles \texttt{render} criteria,
which cover requirements on visual appearance, spatial layout, and motion
that can be verified from rendered views and exported media.
It receives the task context together with three evidence groups: AaR views, original exported media, 
and additional ordered frames sampled from exported videos. 
When visual evidence conflicts, original exported media take precedence
over other views to keep scoring faithful to the
agent's final deliverables. Moreover, a separate blind pass produces
prompt-free descriptions of submission and reference images as auxiliary
grounding evidence. These descriptions supplement the images and are used
only by the visual judge. Unsupported rubric claims receive zero.
\vspace{-5pt}
\paragraph{Dimension Judging.}
To address the limitations of binary rubrics in quantifying continuous perceptual attributes, the VLM additionally evaluates artifact quality along eight dimensions: scene fidelity, physical plausibility, style and color, realism, spatial layout, proportion and scale, lighting and materials, and completeness. Each dimension receives an integer score from 1 to 10 based on the task requirements and available references, with a justification grounded in the visual evidence. Detailed scoring guidelines are provided in Appendix~\ref{app:dimension_scoring}.

\begin{figure}[t]
    \centering
    \includegraphics[width=.9\linewidth]{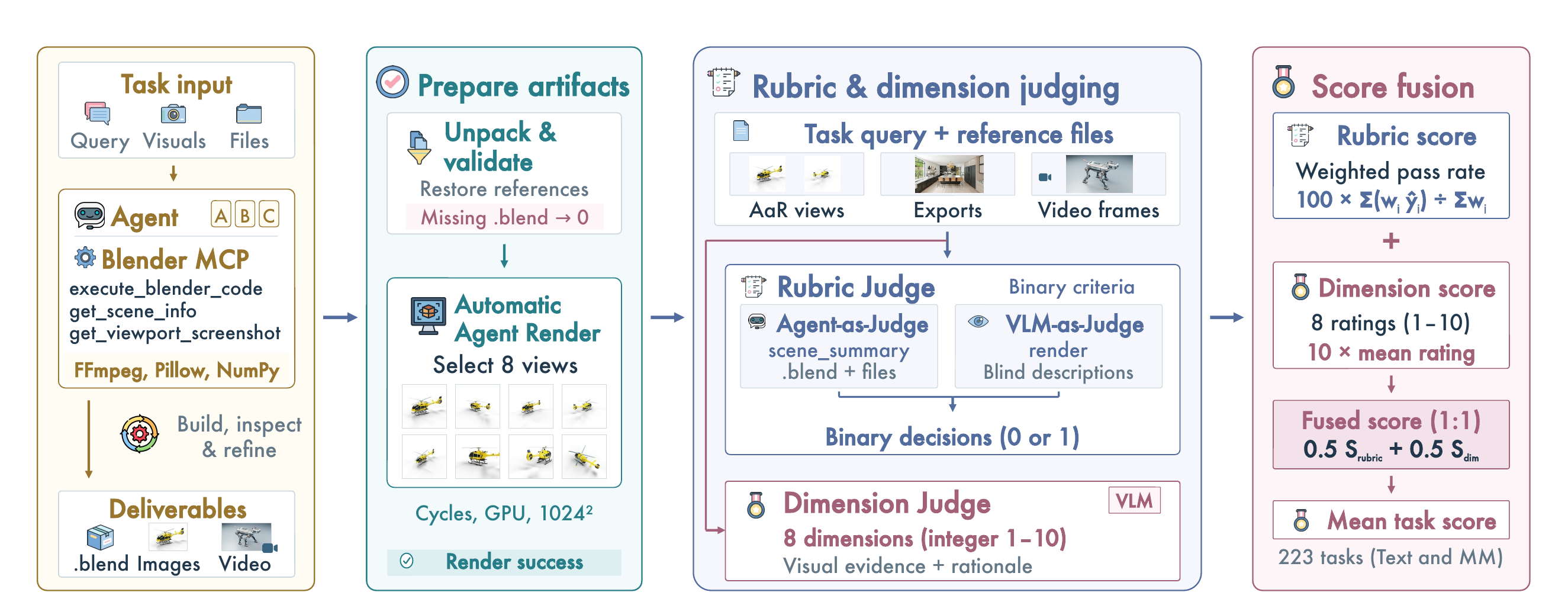}
    \vspace{-5pt}
    \caption{Evaluation pipeline of \benchname{}. Structural and visual evidence support binary rubric decisions, which are fused with dimension scores.}
    \label{fig:eval}
    \vspace{-1em}
\end{figure}
\vspace{-5pt}

\subsection{Metrics}
\label{sec:metrics}

We combine requirement satisfaction and holistic artifact quality into a
single score for each task $q$. Given $m$ binary rubric decisions
$\hat y_j\in\{0,1\}$ with importance weights $w_j$ and $K=8$ dimension
ratings $d_k\in[1,10]$, we compute two scores on a 0--100 scale:
\vspace{-3pt}
\begin{equation}
S_{\mathrm{rubric}}=100\frac{\sum_{j=1}^{m}w_j\hat y_j}
    {\sum_{j=1}^{m}w_j},\qquad
S_{\mathrm{dim}}=100\frac{\sum_{k=1}^{K}d_k}{10K}.
    \label{eq:rubric_score}
\end{equation}
We define the fused task score as $S_{\mathrm{fused}}^{(q)} = \lambda S_{\mathrm{rubric}}^{(q)} + (1-\lambda) S_{\mathrm{dim}}^{(q)}$. We set $\lambda = 0.5$ to place equal emphasis on functional requirement satisfaction and holistic aesthetic quality (a detailed analysis of the $\lambda$ coefficient is provided in Appendix~\ref{app:metric_sensitivity}). We report the mean score over all $Q=223$ tasks, $\mathrm{Score}=\frac{1}{Q}\sum_{q=1}^{Q}S_{\mathrm{fused}}^{(q)}$, together with separate means for the 78 Text and 145 MM tasks. Failed evaluations receive zero and remain in the denominator.

%% file: section/5_experiment.tex
\section{Experiments}
\label{sec:experiments}

\subsection{Evaluation Setup}
\paragraph{Models and environment.}
We compare thirteen agents on the same 223 tasks: 78 Text and 145 multimodal tasks with image or video references. GPT-6-Astra and GPT-5.6-Sol use Codex, while all other agents use Claude Code. Each agent is tested once per task, operating at its highest available reasoning intensity. Agents work in isolated Linux containers, each equipped with one NVIDIA GPU and a pre-installed Linux version of Blender LTS. Crucially, artifact generation and evaluation are executed in strictly separated containers; the evaluation rubrics remain completely invisible to the agents during the generation phase to prevent rubric hacking. Shell, file, image-viewing, and Blender MCP tools support scene creation, rendering, inspection, and revision. Appendix~\ref{app:environment_tools} details the tools and runtime settings.

\paragraph{Evaluation metrics.}
Following the scoring protocol in Section~\ref{sec:evaluation}, we report
mean scores on text-only tasks (Text), multimodal tasks (MM), and all
223 tasks (Score). Rubric / Dimension reports the two component means
on a 0--100 scale, measuring weighted requirement satisfaction and overall
visual quality. Perfect is the percentage of tasks passing every
rubric criterion. Submissions without a \texttt{.blend} file automatically
receive a score of zero because they cannot be rendered for evaluation
and do not meet the task's delivery requirements. For generation
trajectories, we report mean turns, mean output tokens, and the
mean number of tool calls per task. To balance computational cost and evaluation accuracy, we employ \textit{GPT-5.6-terra} for automatic agent rendering, rubric judging, and dimension judging. A detailed analysis of the judge model is provided in Appendix~\ref{app:human_align}.

\subsection{Main Results}
\label{sec:main_results}

\textbf{GPT-6-Astra leads across all completion metrics.}
It achieves an overall Score of 79.9, ahead of Opus-5 at 73.0
(Table~\ref{tab:main223}).
Its lead over Opus-5 is larger on multimodal tasks
(80.7 vs.\ 72.1) than on text-only tasks (78.6 vs.\ 74.8).
GPT-6-Astra also ranks first in both Rubric and Dimension, indicating
advantages in both requirement satisfaction and overall visual quality.
\textbf{However, satisfying every rubric criterion remains challenging.}
GPT-6-Astra achieves a Perfect rate of 42.2\%, compared with 25.6\% for
Opus-5 and at most 15.7\% for the remaining agents.
Thus, even the highest-scoring agents fully satisfy the rubric
on fewer than half of the benchmark tasks.

GPT-5.6-Sol matches DeepSeek-V4.1-Flash on multimodal tasks and achieves a slightly
higher overall Score, yet their rollout statistics differ substantially.
GPT-5.6-Sol averages 39.9K output tokens and 44.0 tool calls per task,
compared with 209K tokens and 168 calls for DeepSeek-V4.1-Flash.
The difference in tool use includes GPT-5.6-Sol's fewer Blender executions and
image view tool calls (Figure~\ref{fig:tool_distribution}).
GPT-6-Astra reinforces this pattern: it achieves the highest overall Score
while using fewer output tokens and tool calls than both Opus-5 and
DeepSeek-V4.1-Flash.
\textbf{These patterns suggest different balances between planning
and iterative tool use.}
The GPT Series Models may favor more extensive planning before acting, whereas DeepSeek-V4.1-Flash relies more on repeated execution and inspection to refine its scenes. Further trajectory analysis could further clarify how planning and iterative refinement contribute to scene quality and interaction efficiency.

\input{tabular/main_exp}

\subsection{Analysis}
\label{sec:experiment_analysis}

\subsubsection{Performance across Task Categories}
\label{sec:task_profiles}

\begin{figure}[t]
    \centering
    \setlength{\abovecaptionskip}{4pt}
    \includegraphics[width=.9\linewidth]{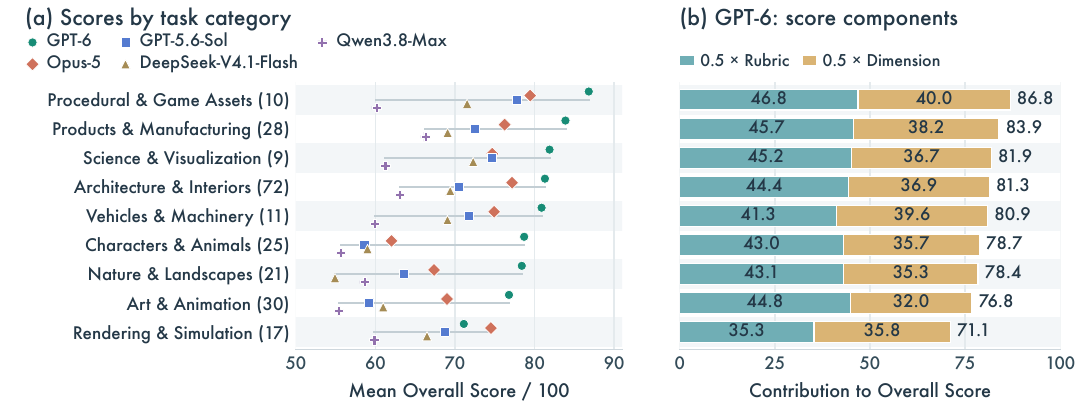}
    \caption{\textbf{Category performance and GPT-6-Astra's score composition.}
    (a) Mean Overall Score for the five highest-scoring agents.
    (b) GPT-6-Astra's mean Rubric and Dimension contributions.}
    \label{fig:category_scores}
\end{figure}

Among the five agents shown in Figure~\ref{fig:category_scores}a,
GPT-6-Astra leads eight of nine categories, scoring highest on Procedural
\& Game Assets (86.8) and Products \& Manufacturing (83.9).
Its largest lead over Opus-5, however, occurs in Characters \& Animals,
where its own score is lower than its overall score (78.7 vs 79.9). 
The category rankings also contain reversals.
Opus-5 outperforms GPT-6-Astra on Rendering \& Simulation, while
Qwen3.8-Max exceeds DeepSeek-V4.1-Flash on Nature \& Landscapes
despite ranking below it overall (Table~\ref{tab:main223}).
These results suggest that modeling strengths vary across agents,
with a shared tendency toward higher scores on procedural assets,
manufactured objects, and architectural scenes.
Natural scenes and characters remain more challenging.

\textbf{GPT-6-Astra's weaker categories reveal distinct trade-offs between visual quality and requirement satisfaction.}
In \textit{Art \& Animation}, the model maintains a high Rubric score (89.6) but suffers the lowest Dimension score (64).
Conversely, \textit{Rendering \& Simulation} yields the lowest Rubric score (70.6) despite achieving a higher Dimension score (71.6), causing it to rank last.
These findings indicate that while GPT-6-Astra reliably fulfills explicit requirements in \textit{Art \& Animation}, it struggles with visual quality. In contrast, for \textit{Rendering \& Simulation}, the bottleneck lies in faithfully reproducing the required behaviors, as exemplified by the failure case in Figure~\ref{fig:failure_modes}a.

\begin{figure}[t]
\centering
\setlength{\abovecaptionskip}{4pt}
\includegraphics[width=.9\linewidth]{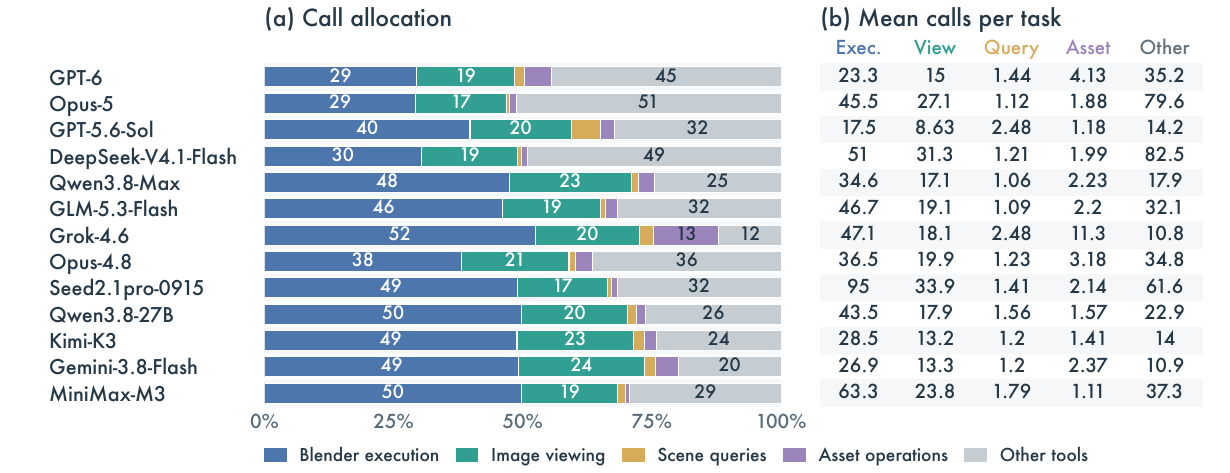}
\caption{\textbf{Tool allocation and interaction volume.}
(a) Distribution of five exclusive tool invocation categories
(b) Mean calls per task. }
\label{fig:tool_distribution}
\end{figure}

\subsubsection{Tool Use}
\label{sec:generation_resources}

\textbf{Tool Allocation Discrepancy vs. Interaction Volume.}
As decoupled in Figure~\ref{fig:tool_distribution}, examining proportional tool allocation alongside absolute volume reveals fundamentally different agent strategies. While top performers like \textbf{GPT-6-Astra} and \textbf{Opus-5} allocate similar budget fractions to \textit{Blender execution} and \textit{Image viewing}, their absolute workloads diverge substantially: \textbf{Opus-5} requires roughly twice as many executions (45.5 vs. 23.3) and views (27.1 vs. 15.0) per task, alongside an inflated volume of auxiliary operations. In contrast, \textbf{Grok-4.6} uniquely leads in \textit{Asset operations} (13\%), demonstrating a new paradigm by prioritizing asset manipulation.

\textbf{Passive Inspection vs. Effective Corrective Feedback.}
Conversely, higher visual call volume does not inherently guarantee effective artifact refinement. Models such as \textbf{Seed2.1pro-0915} and \textbf{DeepSeek-V4.1-Flash} make the most frequent visual queries across all agents (33.9 and 31.3 calls per task), yet their multimodal scores remain moderate or poor. In stark contrast, \textbf{GPT-6-Astra} achieves the top multimodal score (80.7) and rubric completion (87.3) with merely 15.0 image inspections. \textbf{Viewing volume captures the opportunity for feedback without ensuring that the agent can diagnose defects or enact valid corrections.} These aggregate statistics show that more visual inspections do not necessarily coincide with higher scores. They do not establish how effectively each agent uses visual feedback.

\subsubsection{Failure Analysis}
\label{sec:quality_diagnostics}

\begin{figure}[t]
\centering
\setlength{\abovecaptionskip}{3pt}
\includegraphics[width=\linewidth]{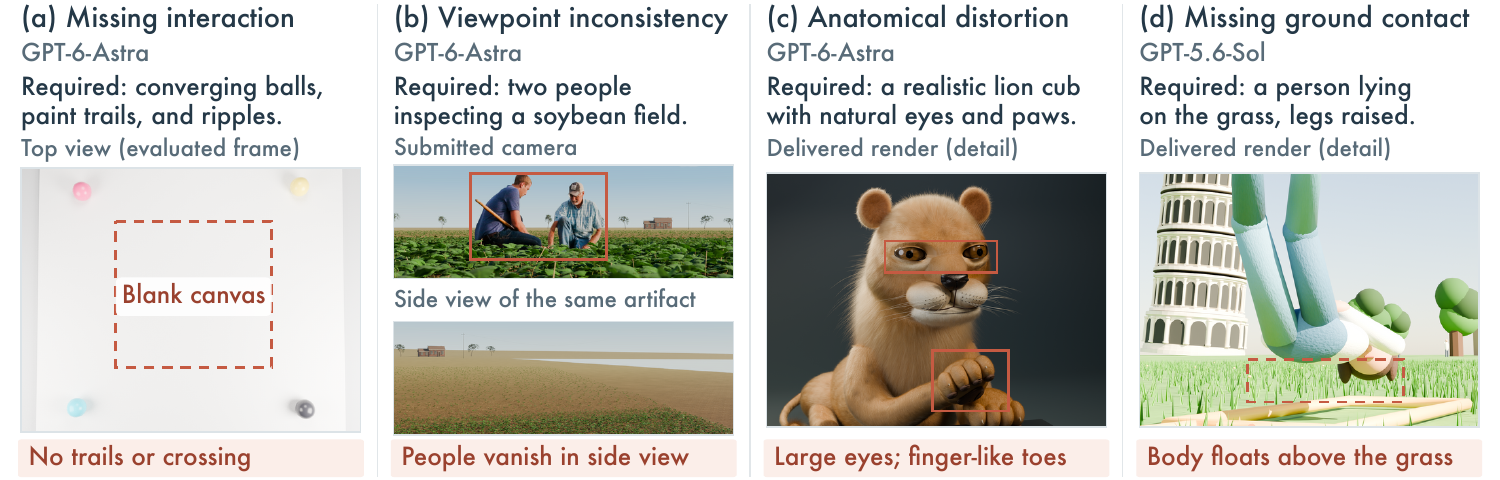}
\caption{\textbf{Example of failure modes.}
(a) Missing paint trails and central crossing, (b) people absent in a side
view, (c) distorted lion-cub anatomy, (d) a reclining person floating above
the lawn.}
\label{fig:failure_modes}
\end{figure}

\textbf{Failure Analysis and Bottlenecks.}
A qualitative audit of failure modes (Figure~\ref{fig:failure_modes}) reveals four recurring failure mechanisms across agents.
(1) \textbf{Missing Dynamic Interactions}: In Figure~\ref{fig:failure_modes}a, GPT-6-Astra satisfies object presence but fails to generate physical interactions, leaving the canvas blank without converging trails.
(2) \textbf{Viewpoint Inconsistency}: Figure~\ref{fig:failure_modes}b demonstrates a lack of true 3D spatial coherence, where people visible in the submitted camera vanish entirely under alternate viewing angles, revealing brittle, view-dependent shortcuts.
(3) \textbf{Fine-grained Geometric Distortions}: The generated lion cub exhibits unnatural, finger-like paws and distorted eyes, reflecting the model's inability to inspect and refine high-frequency anatomical details.
(4) \textbf{Contact Constraint Violations}: The character visibly floats above the lawn, underscoring weak spatial construction and physical simulation ability.
Together, these patterns show that current agents tend to rely on 2D visual shortcuts to make things look right at a glance, but they still struggle to build genuinely solid and physically sound 3D scenes. More thorough failure analysis is provided in Appendix~\ref{app:failure_analysis}.

%% file: tabular/main_exp.tex
\begin{table*}[t]
\centering
\caption{\textbf{Performance on all 223 \benchname{} tasks.} Models are ranked by mean fused Score; bold marks the best completion values. }
\label{tab:main223}
\footnotesize
\setlength{\tabcolsep}{3pt}
\renewcommand{\arraystretch}{1.1}
\begin{lrbox}{\bwmaintablebox}
\begin{tabular}{@{}l rrcrc B ccc@{}}
\toprule
\multirow{2}{*}{\textbf{Model}} & \multicolumn{5}{cB}{\textbf{Completion}} & \multicolumn{3}{c}{\textbf{Rollout statistics}} \\
\cmidrule(lr){2-6}\cmidrule(l){7-9}
 & Text & MM & Score & \shortstack{Rubric /\\Dimension} & \shortstack{Perfect\\(\%)} & Turns & \shortstack{Output\\tokens (K)} & \shortstack{Tool\\calls} \\
\midrule
\bwmodel{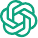}{GPT-6-Astra} & \textbf{78.6} & \textbf{80.7} & \bwtablebest{bwscorewash}{79.9}{0.79930680} & \textbf{87.3} / \textbf{72.6} & \textbf{42.2} & 45.6 & \bwtablebar{bwtokenwash}{84.5}{0.35029720} & 79.1 \\
\bwmodel{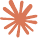}{Opus-5} & 74.8 & 72.1 & \bwtablebar{bwscorewash}{73.0}{0.73027610} & 80.9 / 65.1 & 25.6 & 158 & \bwtablebar{bwtokenwash}{241}{1.00000000} & 155 \\
\bwmodel{openai}{GPT-5.6-Sol} & 71.7 & 65.5 & \bwtablebar{bwscorewash}{67.7}{0.67669316} & 74.7 / 60.7 & 15.7 & 44.9 & \bwtablebar{bwtokenwash}{39.9}{0.16528557} & 44.0 \\
\bwmodel{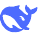}{DeepSeek-V4.1-Flash} & 66.0 & 65.5 & \bwtablebar{bwscorewash}{65.7}{0.65676524} & 72.0 / 59.4 & 11.7 & 171 & \bwtablebar{bwtokenwash}{209}{0.86412437} & 168 \\
\bwmodel{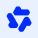}{Qwen3.8-Max} & 59.6 & 61.1 & \bwtablebar{bwscorewash}{60.6}{0.60605969} & 65.9 / 55.3 & 10.3 & 75.1 & \bwtablebar{bwtokenwash}{90.6}{0.37567429} & 72.8 \\
\bwmodel{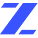}{GLM-5.3-Flash} & 63.1 & 57.8 & \bwtablebar{bwscorewash}{59.7}{0.59674370} & 64.8 / 54.4 & 6.28 & 102 & \bwtablebar{bwtokenwash}{145}{0.60153480} & 101 \\
\bwmodel{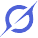}{Grok-4.6} & 60.0 & 55.4 & \bwtablebar{bwscorewash}{57.0}{0.57040397} & 61.4 / 52.7 & 6.73 & 93.4 & \bwtablebar{bwtokenwash}{120}{0.49572524} & 89.8 \\
\bwmodel{claude}{Opus-4.8} & 58.6 & 56.1 & \bwtablebar{bwscorewash}{57.0}{0.56975801} & 60.2 / 53.7 & 5.38 & 97.7 & \bwtablebar{bwtokenwash}{164}{0.67943956} & 95.6 \\
\bwmodel{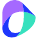}{Seed2.1pro-0915} & 54.7 & 49.5 & \bwtablebar{bwscorewash}{51.3}{0.51307822} & 54.9 / 47.7 & 3.59 & 196 & \bwtablebar{bwtokenwash}{122}{0.50457487} & 194 \\
\bwmodel{qwen}{Qwen3.8-27B} & 46.1 & 48.6 & \bwtablebar{bwscorewash}{47.7}{0.47744489} & 50.0 / 45.5 & 1.35 & 90.0 & \bwtablebar{bwtokenwash}{112}{0.46341592} & 87.4 \\
\bwmodel{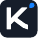}{Kimi-K3} & 41.3 & 49.4 & \bwtablebar{bwscorewash}{46.5}{0.46540346} & 47.9 / 45.2 & 1.79 & 60.5 & \bwtablebar{bwtokenwash}{54.3}{0.22497495} & 58.3 \\
\bwmodel{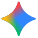}{Gemini-3.8-Flash} & 35.9 & 33.2 & \bwtablebar{bwscorewash}{34.2}{0.34164918} & 29.8 / 38.5 & 0.45 & 56.6 & \bwtablebar{bwtokenwash}{35.9}{0.14891333} & 54.5 \\
\bwmodel{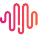}{MiniMax-M3} & 34.1 & 30.6 & \bwtablebar{bwscorewash}{31.8}{0.31827154} & 28.9 / 34.7 & 0.45 & 132 & \bwtablebar{bwtokenwash}{72.2}{0.29937110} & 127 \\
\bottomrule
\end{tabular}
\end{lrbox}
\typeout{BW_MAIN_TABLE_WIDTH=\the\wd\bwmaintablebox; TEXTWIDTH=\the\textwidth}
\resizebox{\textwidth}{!}{\usebox{\bwmaintablebox}}
\par\vspace{4pt}
\vspace{-1em}
\end{table*}

%% file: section/6_conclusion.tex
\section{Conclusion}

We introduced \benchname{}, a benchmark for evaluating visual agents on practical 3D reconstruction, modification, and design in Blender. Its 223 tasks span text-only, image-conditioned, and video-conditioned requests across 9 categories and 16 application scenarios, requiring editable projects and task-specific deliverables. Our evaluation combines task-conditioned multi-view rendering, scene inspection, 2,651 binary criteria, and visual quality assessment to measure both requirement satisfaction and artifact quality. Across 13 agents, GPT-6-Astra achieves the highest overall score of 79.9, yet passes every rubric criterion on only 42.2\% of tasks. Qualitative analyses expose incomplete interactions, inconsistencies across views, geometric distortions, and contact violations, highlighting the challenge of translating task requirements and visual references into coherent 3D structure and behavior. \benchname{} provides a basis for comparing complete agents and diagnosing these gaps, supporting future work on agents that verify and refine their artifacts against both visual evidence and explicit constraints.

%% file: section/7_future.tex
\section{Future Work}
\label{sec:future_work}

We plan to extend BlenderWorld in five directions, covering evaluation,
task design, and experimental scope.

\paragraph{Interactive and Parametric Evaluation.}
We plan to give the agent judge Computer-Use Agent (CUA) tools to interact
with \texttt{.blend} projects and check rigs, animations, and simulations.
By manipulating specified controls and inspecting the resulting states,
the judge could test whether mechanisms behave as intended. We will pair
structured parameter extraction with deterministic scripts for reproducible
checks of dimensions, tolerances, and geometric constraints. For
precision-critical tasks, reference solutions and geometric metrics could
supplement rubric-based judgments. For example, voxel
intersection-over-union would be computed using consistent alignment and
scale conventions.

\paragraph{Higher Complexity and Task Diversification.}
We will expand the benchmark to cover more categories and application
scenarios, with an emphasis on demanding professional workflows. Planned
tasks include localized editing, mesh fault diagnosis and repair, topology
optimization, and retopology, alongside scene creation from scratch. We
also plan to introduce engineering drawings and dimensional tolerances as
richer specifications, and tasks that require revisions to previously
delivered projects. These tasks will assess whether agents preserve
existing content, satisfy interacting constraints, and keep projects
editable as they revise them.

\paragraph{Expert-Aligned and Fine-Grained Rubrics.}
We will work with professional 3D artists to iteratively refine the rubrics
to reflect practical acceptance standards, with particular attention to
modeling reliability, geometric stability, structural consistency, and
production readiness. For organic subjects and other cases in which binary
checks offer limited discrimination, we will explore graded assessments
of individual parts against reference materials. Expert review and
agreement studies will guide rubric calibration and assess consistency
across viewpoints, animation frames, and application domains.

\paragraph{Multi-DCC Ecosystem and Asset Integration.}
We aim to support other mainstream digital content creation (DCC)
applications, including Maya and 3ds Max, as well as additional Blender
versions and MCP tools. Shared task specifications and comparable checks
of submitted artifacts will support evaluation across applications. We
also plan to integrate external asset repositories, such as the Mixamo
motion library, to study workflows that retrieve, adapt, and compose
existing assets. Evaluating settings with and without access to these
repositories will help assess how external assets affect performance.

\paragraph{Broader Model, Reasoning, Harness, and Platform Coverage.}
We plan to evaluate more models across supported thinking-effort settings
and agent harnesses. Within-model studies will vary reasoning effort under
a fixed harness, or compare harnesses with reasoning settings and tool
access kept comparable where supported. Experiments will run under
documented resource budgets and report task quality together with token
usage, latency, tool calls, and cost. Repeated runs on representative
configurations will estimate variability and assess the stability of
rankings. We will also extend the current Linux-based evaluation to Windows
and macOS, and explore providing models with CUA tools for operating
Blender through its graphical interface. This would let us observe how
models navigate the viewport, manipulate objects, and use visual feedback
to inspect and revise their work. Comparisons with API/MCP-based workflows
will examine how task performance, interaction efficiency, and failure
modes differ across interfaces and operating systems.

%% file: section/99_appendix.tex
\clearpage
\appendix

\section{Task Coverage}
\label{app:application_contexts}

Each task has one application-scenario label describing its intended use.
Table~\ref{tab:application_contexts} completes the distribution in
Figure~\ref{fig:blenderworld_statistics}a, including the five smaller
scenarios with ten tasks in total.

\begin{table}[htbp]
\centering
\caption{All 16 application scenarios. Counts sum to 223 tasks.}
\label{tab:application_contexts}
\small
\setlength{\tabcolsep}{5pt}
\begin{tabular}{@{}lr@{\hspace{22pt}}lr@{}}
\toprule
Application scenario & Tasks & Application scenario & Tasks \\
\midrule
Storyboarding / previsualization & 35 & Safety training & 15 \\
Product launches & 21 & Exhibition planning & 15 \\
Game assets & 21 & Construction reporting & 15 \\
E-commerce visuals & 20 & Creative photography & 3 \\
Real-estate marketing & 19 & Interactive exhibits & 3 \\
Scientific illustration & 18 & Travel promotion & 2 \\
Renovation proposals & 18 & Animation / rigging & 1 \\
Classroom teaching & 16 & AR avatars & 1 \\
\bottomrule
\end{tabular}
\end{table}

The input combinations are 56 briefs without references or attachments,
22 with specifications only, 93 with images only, 37 with images and
specifications, and 15 with videos. Thus, Text denotes the 78 tasks
without image or video references, including 22 with specifications;
Multimodal denotes the other 145 tasks. The 59 specification attachments comprise
28 Markdown, 20 YAML, and 11 JSON files. \input{section/appendix_rubric_generation}

\section{Scoring Details}
\label{app:dimension_scoring}

\paragraph{Dimension definitions and anchors.}
Table~\ref{tab:dimension_guidelines} summarizes the eight dimensions in the
judge instructions. Each receives an integer rating using the same anchors:
\textbf{1--2}, absent, broken, or unrecognizable;
\textbf{3--4}, major failures and only loose resemblance;
\textbf{5--6}, acceptable with clear defects;
\textbf{7--8}, mostly correct with minor issues; and
\textbf{9--10}, essentially faithful with only trivial flaws.
The judge cites decisive files or views and records the observed issues
for each rating. The brief and references define the target, including
any intended stylization.

\begin{table}[htbp]
\centering
\caption{Dimension-specific assessment. These visual ratings complement
the binary checks of task requirements and native project properties.}
\label{tab:dimension_guidelines}
\small
\renewcommand{\arraystretch}{1.14}
\begin{tabular}{@{}p{0.24\linewidth}p{0.72\linewidth}@{}}
\toprule
Dimension & Assessment focus \\
\midrule
Scene fidelity & Recognizable target scene, principal subjects, and overall arrangement. \\
Physical plausibility & Support, attachment, thickness, coherent structures, and absence of floating or intersecting parts. \\
Style and color & Recognizable object shapes and silhouettes, with colors appropriate to the target. \\
Realism & A believable, coherent image with adequate contrast and render quality; appropriate simplification may preserve object identity. \\
Spatial layout & Relative positions, orientations, and spacing, including adjacency and front--back relationships. \\
Proportion and scale & Plausible size ratios between objects, their parts, and the surrounding scene. \\
Lighting and materials & Surface appearance, illumination, and shadows appropriate to the intended result; suitable matte finishes are acceptable. \\
Completeness & Presence of the required objects in approximately the right counts, without important omissions. \\
\bottomrule
\end{tabular}
\end{table}

\section{Runtime and Resource Accounting}
\label{app:environment_tools}

\subsection{Runtime Configuration}
The reference runtime uses Blender 4.2.10 LTS, Codex CLI 0.147.0,
Claude Code 2.1.220, and Node.js 22.17.0. GPT-6-Astra and GPT-5.6-Sol use
Codex; the other eleven systems use Claude Code. MiniMax-M3's 48-task
batch uses Claude Code 2.1.241. Isolated Linux workspaces provide one
NVIDIA GPU per task, a virtual display, and Cycles rendering through
OptiX or CUDA. The final workspace is archived for evaluation.

\subsection{Available Tools}
The harness provides shell execution, file operations, and local image
viewing. Agents can write scripts, launch Blender in background mode,
inspect renders, and revise their projects. FFmpeg and ImageMagick support
media processing, alongside NumPy, Pillow, SciPy, and ImageIO.
The Blender MCP endpoints below omit the common
\texttt{mcp\_\_blender\_\_} prefix.

\begin{itemize}[leftmargin=*,itemsep=4pt,topsep=4pt]
\raggedright
\item \textbf{Scene editing and inspection.}
\texttt{execute\_blender\_code} runs Python inside Blender.
\texttt{get\_scene\_info} and \texttt{get\_object\_info} return scene
structure and object properties; \texttt{get\_viewport\_screenshot}
provides visual feedback.

\item \textbf{Materials and environment assets.}
\texttt{get\_polyhaven\_status}, \texttt{get\_polyhaven\_categories},
\texttt{search\_polyhaven\_assets}, and
\texttt{download\_polyhaven\_asset} support availability checks,
discovery, and retrieval. \texttt{set\_texture} applies textures.

\item \textbf{External model retrieval.}
\texttt{get\_sketchfab\_status}, \texttt{search\_sketchfab\_models},
\texttt{get\_sketchfab\_model\_preview}, and
\texttt{download\_sketchfab\_model} provide availability checks,
search, previews, and model import.

\end{itemize}

\subsection{Resource Accounting}
\label{app:process_analysis}

We record resource counters (including output tokens, interaction turns, and tool calls) across the full suite of 223 tasks for all evaluated agents. These counters incorporate unsuccessful calls. Output tokens reflect provider-reported completion counts. A turn is uniformly defined as a single model-response round within the agent runtime. 

As shown in Figure~\ref{fig:tool_distribution}, tool calls are grouped into five exclusive categories: \emph{Blender execution} (direct Blender/Python and MCP code calls); \emph{image viewing} (reference images, renders, and viewport screenshots); \emph{scene queries} (scene/object information); \emph{asset operations} (discovery, status, retrieval, generation, import, and texture assignment); and \emph{other}. Compound calls are counted as a single operation. Category means are computed by dividing pooled counts by the total 223 tasks, and shares are calculated against total tool calls. These metrics strictly describe requested operations, rather than guaranteed successful executions or matched compute costs.

\input{section/appendix_experiments}

%% file: section/appendix_rubric_generation.tex
\section{Rubric Construction}
\label{app:rubric_generation}

\paragraph{From sources to acceptance conditions.}
The generator reads the query, actual references, and full attachments,
then records binding requirements with source locations and verification
methods. This ledger covers structure, space, appearance, function,
presentation, and delivery as applicable. Requirements are extracted before
compression, preserving both explicit constraints and allowed design freedom.

Related requirements form a coherent acceptance unit, such as a component,
interface, functional outcome, or deliverable. Each criterion records its
identifier, subject, source, category, importance, evidence type, and
self-contained pass/fail conditions. A merged criterion passes only when
\emph{all} constituent conditions hold; its fail condition is the logical
complement. Major, secondary, and detail criteria receive weights 3, 2,
and 1.

\paragraph{Category vocabulary.}
Each criterion receives one of the 13 categories in
Table~\ref{tab:rubric_categories}, according to its principal focus.
Its full acceptance conditions retain any requirements spanning several
aspects. The six ledger dimensions guide source coverage; these categories
describe the resulting criteria. The \texttt{holistic} category is a
task-specific binary check, separate from the eight quality ratings in
Appendix~\ref{app:dimension_scoring}.

\begin{table}[htbp]
\centering
\caption{Category vocabulary for binary rubrics. Properties are checked
when required by the task materials.}
\label{tab:rubric_categories}
\small
\setlength{\tabcolsep}{6pt}
\renewcommand{\arraystretch}{1.10}
\begin{tabular}{@{}p{0.21\linewidth}p{0.75\linewidth}@{}}
\toprule
Category & Principal focus \\
\midrule
\texttt{presence} & Required subjects or components and their recognizable identity. \\
\texttt{count} & Number of objects, parts, or repeated units. \\
\texttt{geometry} & Shape, silhouette, topology, modeling details, and structural completeness. \\
\texttt{scale} & Absolute dimensions and relative proportions. \\
\texttt{layout} & Placement, arrangement, and pose of objects or components. \\
\texttt{space\_layout} & Scene organization, zoning, clearances, and circulation. \\
\texttt{material} & Surface identity, texture, roughness, transparency, and reflection. \\
\texttt{color} & Specified colors, palettes, and their distribution. \\
\texttt{lighting} & Light configuration, illumination effects, shadows, and brightness. \\
\texttt{composition} & Viewpoint, framing, subject coverage, occlusion, and visual readability. \\
\texttt{physics} & Contact, support, collision, deformation, and motion. \\
\texttt{function} & Functional outcomes, editability, engineering settings, and file delivery. \\
\texttt{holistic} & Task-specific overall identity, style, and completeness. \\
\bottomrule
\end{tabular}
\end{table}

\paragraph{Assigning observable evidence.}
Rendered views and exported media support visible shape, appearance,
composition, and motion. Scene/file inspection supports exact dimensions,
topology, internal structure, configuration, and output metadata. A
criterion combining native properties and visible effects uses the
scene-inspection route, whose judge can also consult images. The frozen
set assigns 1,107 criteria to \texttt{render} and 1,544 to
\texttt{scene\_summary}.

\paragraph{Pilot diagnosis and independent review.}
GPT-5.6-Sol drafts and edits the rubrics. Pilot judgments of its outputs
flag ambiguity, repeated decisions, and missing evidence. Review traces
each issue to criterion content, evidence collection, or judge execution;
corrections require support from the task sources. At least two models
independently inspect the same candidate and materials without seeing
other reviews or evaluated-model scores. Reviewers include GPT-5.6-Sol, Gemini 3.1 Pro Preview, and Claude Opus 4.8.
They check coverage, unsupported constraints, duplication, merging, weights,
pass/fail consistency, and evidence suitability. Human review complements
this process.

\paragraph{Cross-checking and freezing.}
Findings cite sources and a counterexample: a valid output that would
fail, or a violating output that would pass. Reviewers cross-check these
findings, and the editor records the adopted corrections and reasons.
Changes in meaning receive another independent review. Finalization
requires resolution of substantive findings and checks of identifiers,
category/evidence values, weights, binary fields, and source coverage.
Snapshots and review records preserve provenance; accepted conditions,
weights, and evidence assignments are frozen for evaluation.

%% file: section/appendix_experiments.tex
\section{Supplementary Analysis and Examples}
\label{app:experiment_analysis}

\subsection{Detailed Analysis of Judge Model}
\label{app:human_align}

We evaluate the judge models on 150 pairs of artifacts generated for the same input, with each pair assessed by three human reviewers. The majority human preference label, $C_i$, requires at least two of the three reviewers to agree on A, B, or a tie. Three cases where A, B, and a tie each received one vote are excluded, leaving 147 cases for evaluation. For each judge, we examine the Rubric, Dimension, and Overall scores, where the Overall score is the equally weighted average of the Rubric and Dimension scores (i.e., $0.5\text{ Rubric}+0.5\text{ Dimension}$). To calculate agreement, the majority preference label $C_i$ is encoded as $+1$ for preferring A, $-1$ for preferring B, and $0$ for a tie. The machine's predicted preference, $J_i=\text{sign}(s_{iA}-s_{iB})$, must exactly match the majority label $C_i$. Preference correlation metrics (Pearson's $r$, Spearman's $\rho$, and Kendall's $\tau_b$) are derived by computing the correlation between the raw score difference $s_{iA}-s_{iB}$ and the majority label $C_i$ across these 147 cases.

We additionally report pairwise agreement on cases with a small average score gap ($<10$ points). This small-gap subset contains 70 eligible cases satisfying $\vert{}\bar{s}_A-\bar{s}_B\vert{}<10$, where $\bar{s}_X$ is the mean Overall score of artifact $X$ across the three judge models (GPT-5.6-terra, Gemini-3.8-Flash, and DeepSeek-V4.1-Flash).

Table 6 shows that the Overall scores achieve the highest pairwise agreement on the full set of 147 cases for all judge models: 82.99\% for GPT-5.6-terra, 77.55\% for Gemini-3.8-Flash, and 76.87\% for DeepSeek-V4.1-Flash. However, on the small-gap subset, the agreement for these three models decreases to 68.57\%, 58.57\%, and 55.71\%, respectively. This indicates that alignment between machine and human judgments becomes significantly more difficult when the artifacts receive very similar average scores. We also observe that the Dimension scores consistently outperform the Rubric scores in terms of agreement across all models. For all preference correlation metrics, GPT-5.6-terra's Overall score performs the best (Pearson's $r=0.714$, Spearman's $\rho=0.744$, Kendall's $\tau_b=0.607$). \textbf{These findings denote that GPT-5.6-terra is the judge model that best aligns with human scoring preferences. Its scoring is relatively reliable and accurate.}
\input{tabular/Major_voting_three_human_align}

\subsection{Completion across Rubric Thresholds}
\label{app:threshold_coverage}

\begin{figure}[htbp]
\centering
\includegraphics[width=\linewidth]{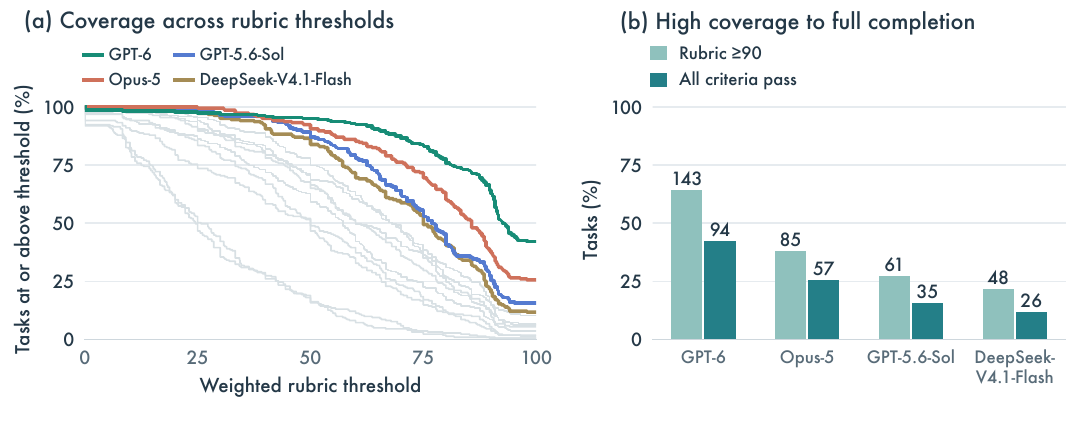}
\caption{\textbf{Coverage at increasing Rubric thresholds.}
(a) Fraction of the 223 tasks at or above each threshold; gray curves
represent the other nine systems. (b) Thresholds 90 and 100 for the four
highest-scoring systems. Bar labels give task counts.}
\label{fig:threshold_coverage}
\end{figure}

Figure~\ref{fig:threshold_coverage} measures the fraction of tasks whose
weighted Rubric score reaches threshold $t$:
$C_m(t)=\frac{1}{223}\sum_{q=1}^{223}\mathbf{1}[R_{mq}\geq t]$.
Missing scores remain zero. Because every criterion has positive weight,
$t=100$ recovers Perfect. GPT-6-Astra covers 143 tasks at $t=90$ but 94 at
$t=100$; Opus-5 drops from 85 to 57. Thus, 49 GPT-6-Astra outputs and 28
Opus-5 outputs satisfy at least 90\% of the criterion weight while still
missing at least one rubric. The curves distinguish high
weighted coverage from complete acceptance on the fixed task set;
only the reporting threshold changes.

\subsection{Component Scores and Unmet Criteria}
\label{app:completion_diagnostics}

\begin{figure}[htbp]
\centering
\includegraphics[width=\linewidth]{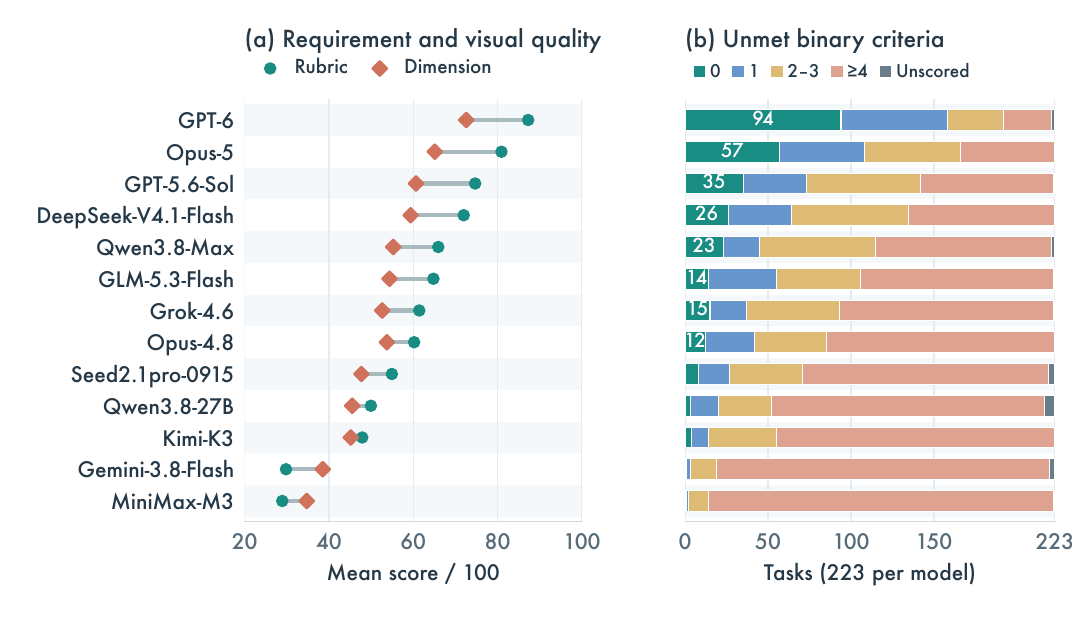}
\caption{\textbf{Score components and unmet requirements.}
(a) Task-mean Rubric and Dimension scores, retaining all 223 tasks with
missing components set to zero. (b) Task counts by number of unmet binary
criteria, with unscored evaluations shown separately. Zero unmet criteria
corresponds to Perfect.}
\label{fig:completion_diagnostics}
\end{figure}

Figure~\ref{fig:completion_diagnostics} places the two component means
alongside the number of unmet criteria per task. GPT-6-Astra and Opus-5 lead
on both components, but their full-pass counts conceal additional
variation: GPT-6-Astra has 94 fully passing outputs and 64 with one unmet
criterion, compared with 57 and 51 for Opus-5. Moreover, one
criterion may combine several conditions, so a single unmet criterion
does not establish that repair is easy. 

\subsection{Overlap in Fully Completed Tasks}
\label{app:paired_completion}

\begin{figure}[htbp]
\centering
\includegraphics[width=\linewidth]{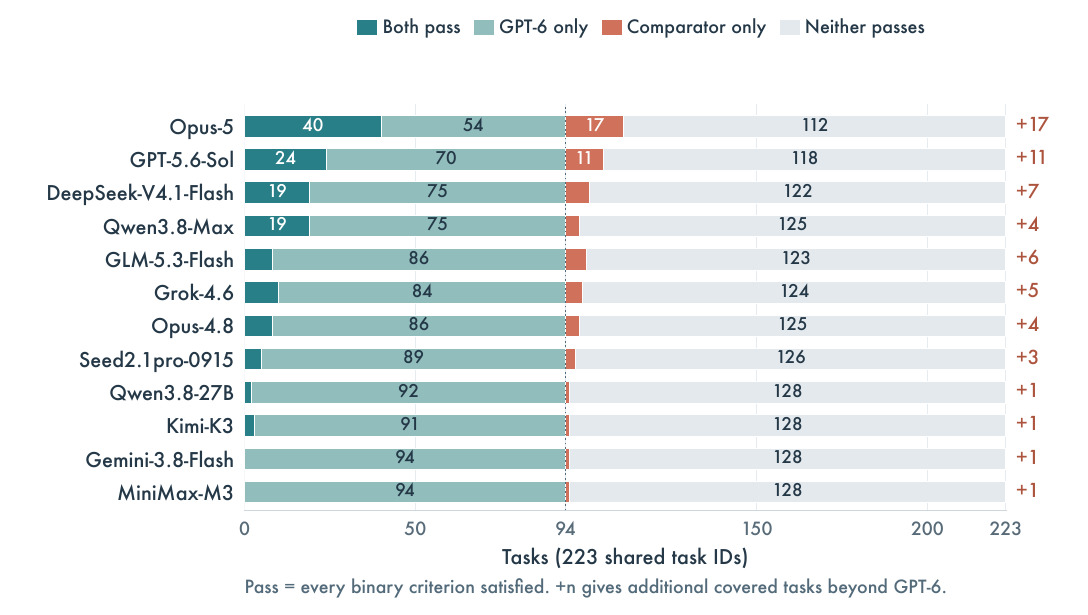}
\caption{\textbf{Paired overlap in complete rubric satisfaction.}
Each bar partitions the same 223 tasks by whether GPT-6-Astra and the comparator
pass every binary criterion. The dotted line marks GPT-6-Astra's 94 full passes;
$+n$ counts comparator-only passes beyond that set.}
\label{fig:paired_completion}
\end{figure}

Figure~\ref{fig:paired_completion} compares each system's fully passing
tasks with GPT-6-Astra's on the same 223 task identifiers. GPT-6-Astra and Opus-5
both pass 40 tasks; 54 pass only under GPT-6-Astra and 17 only under Opus-5,
leaving 112 passed by neither.
GPT-5.6-Sol and DeepSeek-V4.1-Flash contribute 11 and 7 additional tasks,
respectively, beyond GPT-6-Astra. These counts show that the successful task
sets are not fully nested, even the worst Minimax-M3 still has 1 perfect task beyond GPT-6. 

\subsection{Dimension and Rubric Profiles}
\label{app:score_breakdown}

\begin{figure}[htbp]
\centering
\includegraphics[width=\linewidth]{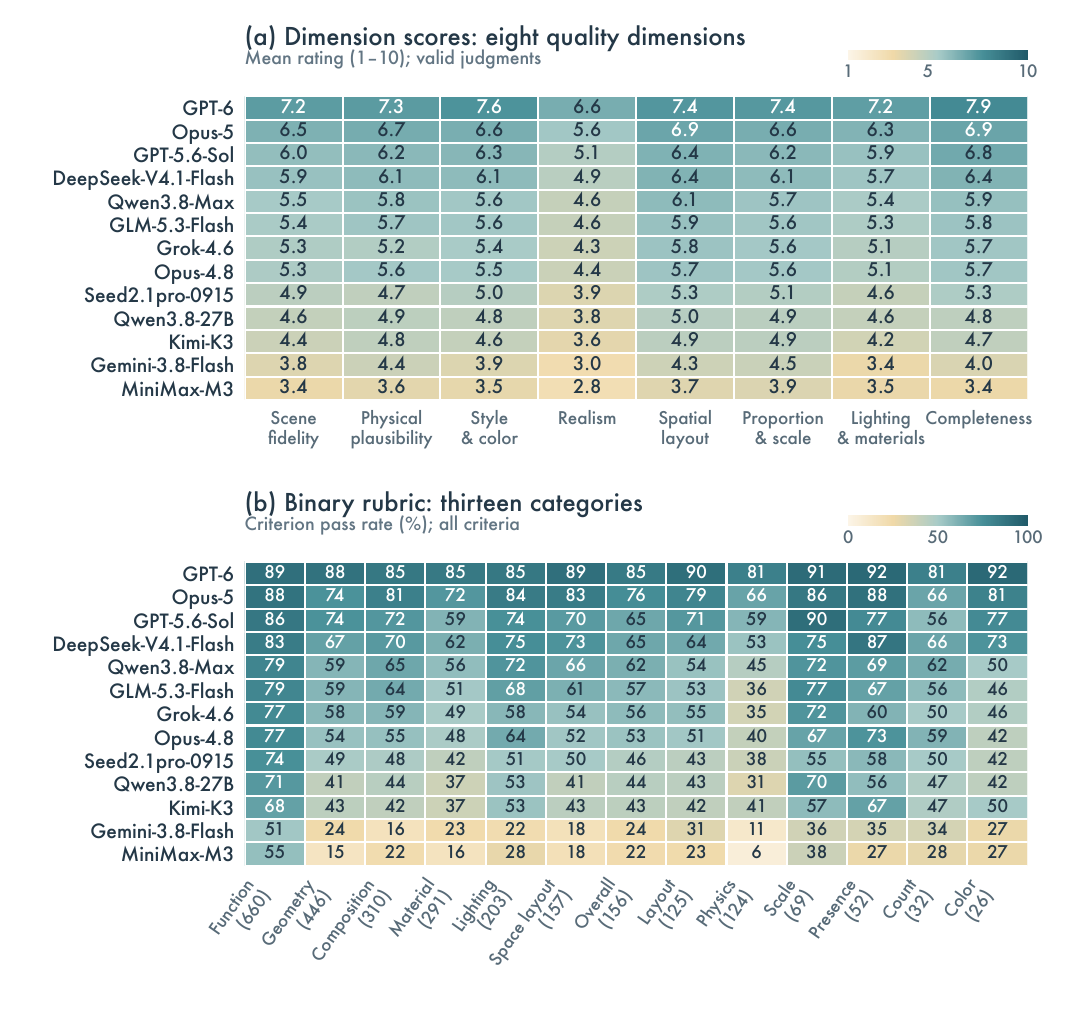}
\caption{\textbf{Quality dimensions and binary-criterion categories.}
(a) Mean 1--10 ratings over valid dimension judgments only.
(b) Unweighted criterion pass rates, counting missing outcomes as non-passes;
Parentheses give criteria per system. }
\label{fig:score_breakdown}
\end{figure}

Figure~\ref{fig:score_breakdown} separates eight quality ratings from
the 13 binary-criterion categories. Realism has the lowest mean dimension
rating for every system. For GPT-6-Astra, it averages 6.61/10, compared with
7.89/10 for completeness; Opus-5 averages 5.62/10 on realism.
The criterion profiles reveal variation in requirement satisfaction:
Opus-5 passes 87.6\% of function criteria, compared with 73.5\% for
geometry, 71.8\% for material, and 66.1\% for physics, while GPT-6-Astra exceeds
80\% in every category. These criterion rates pool different numbers
of checks, shown below the heatmap. They summarize category-specific
outcomes, whereas the main Rubric score normalizes importance weights
within each task.

\subsection{Sensitivity to the Fusion Weight}
\label{app:metric_sensitivity}

\begin{figure}[htbp]
\centering
\includegraphics[width=\linewidth]{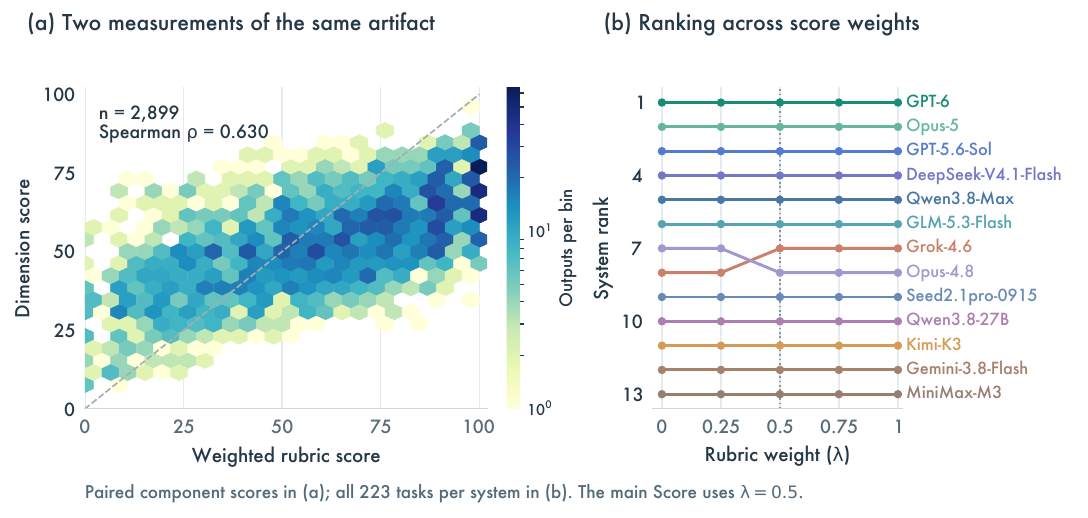}
\caption{\textbf{Score components and fusion-weight sensitivity.}
(a) Paired scores for 2,899 outputs, with logarithmic bin counts.
(b) System ranks at five rubric weights; the main Score uses
$\lambda=0.5$. All 223 tasks remain in each system mean.}
\label{fig:metric_sensitivity}
\end{figure}

We recompute $S=\lambda S_{\mathrm{rubric}}+
(1-\lambda)S_{\mathrm{dim}}$ at
$\lambda\in\{0,0.25,0.5,0.75,1\}$, holding the selected artifacts and
judgments fixed. Each mean retains 223 tasks, unscored evaluations remain
zero. The ranking is unchanged at these weights except for Grok-4.6 and
Opus-4.8: Opus-4.8 ranks seventh at 0 and 0.25, and Grok-4.6 ranks seventh
at 0.5, 0.75, and 1 (Figure~\ref{fig:metric_sensitivity}b).

Although the overall model rankings are highly stable, at a micro level, the two scoring criteria actually show differences when evaluating specific 3D artifacts. As illustrated by the scatter plot in Figure~\ref{fig:metric_sensitivity}a, the Spearman correlation coefficient across the 2,899 output artifacts is 0.630. \textbf{While this indicates a positive overall trend—meaning artifacts that follow instructions better generally tend to have better visual quality, the data points are still widely dispersed rather than tightly hugging the diagonal line.} The underlying reason is that they measure completely different aspects of capability. For instance, some outputs dutifully satisfy all objective instructions (achieving a high Rubric score), but their rendered visual and physical quality is poor, suffering from issues like mesh clipping or unnatural lighting (resulting in a low Dimension score). Conversely, some outputs look visually stunning and realistic, but fail to strictly follow the specific task constraints. Therefore, even though these two measurement methods yield a largely consistent overall model leaderboard, they retain their own distinct and irreplaceable focuses when evaluating individual generated artifacts.

\subsection{Task-Category Profiles for All Systems}
\label{app:domain_profiles}

\begin{figure}[htbp]
\centering
\includegraphics[width=\linewidth]{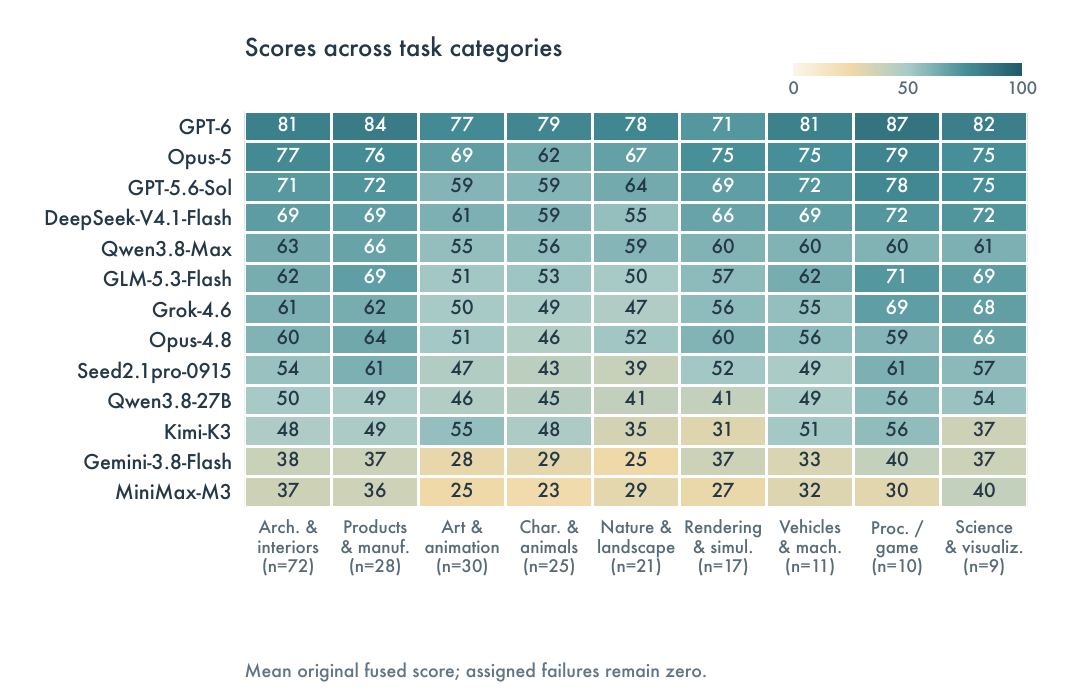}
\caption{\textbf{Task-category scores for all thirteen systems.}
Cells show mean fused Score on each category's fixed task subset, including
assigned zeros. Category sizes appear below the columns; cells are rounded
to the nearest integer.}
\label{fig:domain_profiles}
\end{figure}

Figure~\ref{fig:domain_profiles} extends the five-system comparison in
Figure~\ref{fig:category_scores}a to all thirteen systems. Every system
scores higher on Procedural \& Game Assets than on Art \& Animation or
Characters \& Animals. GPT-6-Astra leads eight categories; Rendering \& Simulation
is the exception, where Opus-5 scores 74.5 versus GPT-6-Astra's 71.1.
GPT-6-Astra's category means range from 71.1 in Rendering \& Simulation to
86.8 in Procedural \& Game Assets.

\begin{figure}[htbp]
\centering
\includegraphics[width=\linewidth]{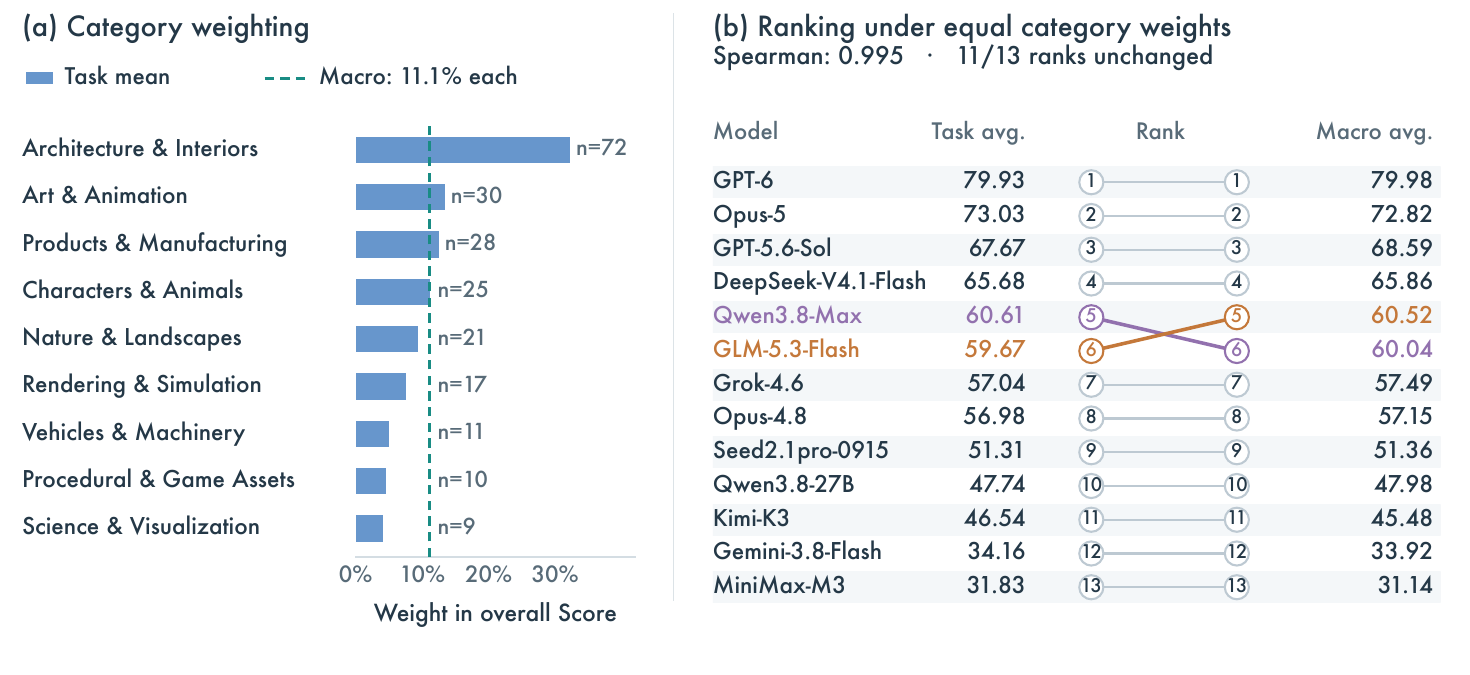}
\caption{\textbf{Rankings are largely stable under equal category weighting.} (a) Bars show each category's weight in the task-level mean; labels give task counts. The dashed line marks the equal 1/9 weight used by the macro-average. (b) Scores and ranks under the two aggregation rules for the same thirteen systems and 223 selected outputs per agent. Colored paths mark the sole rank exchange: Qwen3.8-Max and GLM-5.3-Flash at positions 5 and 6. Scores beside each endpoint follow that side's rank order.}
\label{fig:category_macro_rank}
\end{figure}

\textbf{The leading-system ranking persists under equal category weighting.} Because the nine categories contain 9--72 tasks, we also average their mean Scores with equal weights, keeping the same 223 outputs per system. The macro-average and task-level rankings have Spearman $\rho = 0.995$ and Kendall $\tau = 0.974$: the top four retain their order, and only Qwen3.8-Max and GLM-5.3-Flash exchange fifth and sixth place (Figure~\ref{fig:category_macro_rank}).

\subsection{Connecting Defects to Actions}
\label{app:failure_analysis}

\begin{figure}[htbp]
\centering
\includegraphics[width=\linewidth]{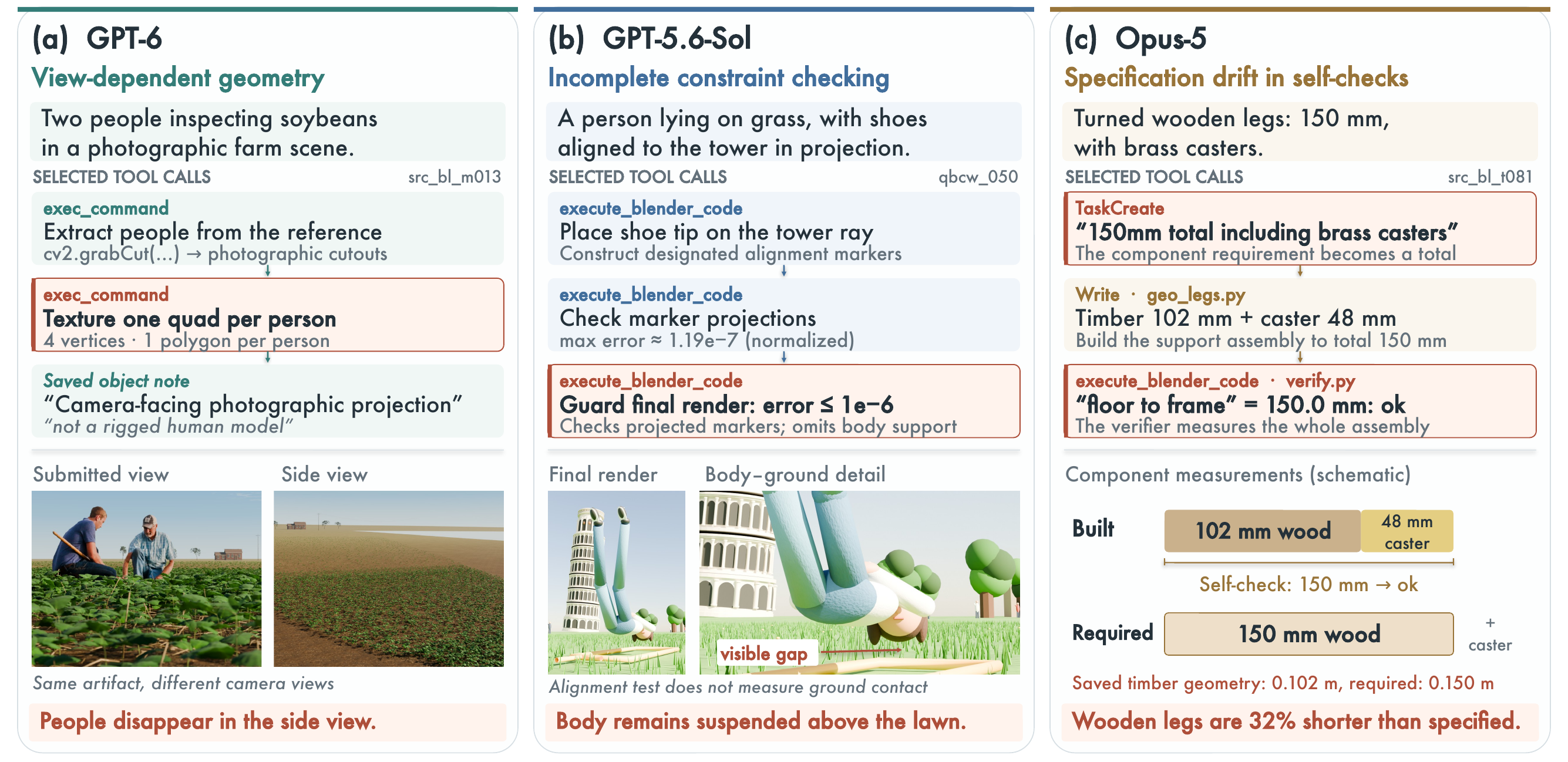}
\caption{\textbf{Selected operations in three failure trajectories.}
(a) GPT-6-Astra maps reference people onto textured flat quads.
(b) GPT-5.6-Sol checks projection markers while the body remains unsupported.
(c) Opus-5 checks the combined leg-and-caster height in place of the
specified timber length. Images in (a) and (b) are archived renders;
the dimension bars in (c) are schematic. }
\label{fig:failure_trajectories}
\end{figure}

Two artifacts in Figure~\ref{fig:failure_modes} are linked to their
selected generation traces with explanations below. Figure~\ref{fig:failure_trajectories}
details selected operations for the \textit{viewpoint inconsistency} and \textit{missing ground contact}
cases, and also adds an Opus-5 example to explain another defect.

\paragraph{Viewpoint inconsistency.}
For this case, GPT-6-Astra segments the two people from the reference
and maps each crop onto a single flat quad, with separate shadow proxies.
The trajectory analysis in the figure~\ref{fig:failure_trajectories} also indicates that the established character object is bound to a camera-facing photographic projection, rather than to a human model with a skeleton. This representation explains their appearance from the submitted camera
and disappearance in the side view. The frozen judgment likewise identifies
one-sided photographic cards. The defect is the missing volume needed
for this scene's alternate views; texture refinement alone cannot supply it.

\paragraph{Missing Ground Contact.}
In this task, the goal was to generate a forced-perspective image featuring a person lying on the grass with their shoes visually aligned with a distant tower. GPT-5.6-Sol demonstrated intelligence and rigor by utilizing code tools to establish alignment markers and precisely calculate projection errors; it even implemented a strict validation step requiring the alignment error to remain below $1\times10^{-6}$. However, the model’s fatal flaw lay in its obsessive focus on satisfying the mathematical constraints of the 2D projection while completely overlooking fundamental 3D physical realities—specifically, the need for the figure to be physically supported by the ground. As shown in the detailed view on the right, although the feet and the tower aligned perfectly from the specific camera angle, the person's body was actually suspended absurdly in mid-air above the grass, leaving a clearly visible gap. This highlights a common limitation in current agents: they tend to solve specific, explicit instructions mechanically but lack a holistic grasp of 3D spatial structures and physical laws such as gravity and contact.

\paragraph{Specification drift in a self-check.}
For this case, the input brief requires 150\,mm wooden legs with
brass casters. Opus-5's \texttt{TaskCreate} instead specifies 150\,mm
including the casters, and its geometry script builds 102\,mm of timber
on 48\,mm casters (Figure~\ref{fig:failure_trajectories}c).
The subsequent verifier measures the floor-to-frame height and reports
150.0\,mm as passing. The frozen scene inspection separately measures
the timber at 102\,mm, 32\% shorter than required, and marks R02 as failed.
The self-check thus confirms the false measurement target while
leaving the original component-length requirement unsatisfied.

\input{section/appendix_case_studies}

%% file: tabular/Major_voting_three_human_align.tex
\begin{table*}[t]
\centering
\caption{\textbf{Alignment of automatic judges with human preferences.} We evaluate 150 pairs of different artifacts generated for the same input, each rated by three human reviewers. Three cases with one vote for each category are excluded, leaving 147 cases. We compare rubric, dimension, and overall scores with the majority human preference.}
\label{tab:human-majority-skip-handling}
\begingroup
\small
\setlength{\tabcolsep}{4.1pt}
\renewcommand{\arraystretch}{1.25}
\resizebox{\textwidth}{!}{%
\begin{tabular}{@{}llcc@{\hspace{12pt}}ccc@{}}
\toprule
\multirow{3}{*}{Judge model} & \multirow{3}{*}{Aspect} & \multicolumn{2}{c}{Pairwise agreement ($\uparrow$)} & \multicolumn{3}{c}{Preference correlation ($\uparrow$)} \\
\cmidrule(lr){3-4}\cmidrule(l){5-7}
& & Overall & Mean score gap $<10$ & Pearson & Spearman & Kendall \\
& & 147 cases & 70 cases & $r$ & $\rho$ & $\tau_b$ \\
\midrule
\multirow{3}{*}{GPT-5.6-terra} & Rubric & 73.47\% & 51.43\% & 0.643 & 0.674 & 0.554 \\
 & Dimension & 79.59\% & 64.29\% & 0.708 & 0.722 & 0.596 \\
 & \textbf{Overall} & 82.99\% & 68.57\% & 0.714 & 0.744 & 0.607 \\
\midrule
\multirow{3}{*}{Gemini-3.8-Flash} & Rubric & 56.46\% & 32.86\% & 0.535 & 0.559 & 0.474 \\
 & Dimension & 76.19\% & 58.57\% & 0.622 & 0.691 & 0.569 \\
 & \textbf{Overall} & 77.55\% & 58.57\% & 0.652 & 0.681 & 0.556 \\
\midrule
\multirow{3}{*}{DeepSeek-V4.1-Flash} & Rubric & 65.99\% & 38.57\% & 0.593 & 0.609 & 0.503 \\
 & Dimension & 74.83\% & 57.14\% & 0.623 & 0.634 & 0.522 \\
 & \textbf{Overall} & 76.87\% & 55.71\% & 0.657 & 0.674 & 0.549 \\
\bottomrule
\end{tabular}%
}
\endgroup
\vspace{3pt}

\end{table*}

%% file: section/appendix_case_studies.tex
\clearpage
\subsection{Paired Case Studies across Six Tasks}
\label{app:paired_case_studies}

\begingroup
\setlength{\abovecaptionskip}{4pt}
\setlength{\belowcaptionskip}{0pt}
\newcommand{\pairedcasepanel}[3]{%
  \par{\raggedright\small\noindent\textbf{#1} #2\par}\vspace{1pt}%
  \includegraphics[page=#3,trim=14bp 42bp 14bp 87bp,clip,height=0.24\textheight]{figures/appendix/six_task_comparisons.pdf}%
  \par\vspace{3pt}}

\begin{figure}[!ht]
\centering
\pairedcasepanel{(a) Exhibition booth.}{Recreate the reference layout, branded panels, and hatch with a visible lunar rover.}{1}
\pairedcasepanel{(b) Lion cub.}{Recreate the central cub with realistic anatomy and distinct fur, whiskers, and nose; deliver three 1080p views.}{2}
\pairedcasepanel{(c) Chesterfield sofa.}{Build a $2200\!\times\!950\!\times\!750$\,mm sofa with 150\,mm wooden legs plus casters; render at 4K.}{3}
\caption{\textbf{Construction and verification across three typical workflows.}
Opus-5 is on the left and GPT-6-Astra on the right. Both booth outputs pass all
binary criteria. Both lion cub pipelines construct fur and rigging, yet their
dimension scores remain 37.50 and 53.75. The sofa reverses the ordering
across components: Opus-5 has higher Dimension (with better lighting), while GPT-6-Astra has higher
Rubric; Figure~\ref{fig:failure_trajectories}c details the timber-length
error.}
\label{fig:paired_cases_appearance}
\end{figure}

\clearpage
\begin{figure}[!ht]
\centering
\pairedcasepanel{(d) Treasure chest.}{Build a chest with a lid, hinges, lock, and riveted trim; bake details onto a mesh of at most 3,000 faces.}{4}
\pairedcasepanel{(e) Cell membrane.}{Illustrate a curved bilayer with inward tails, membrane protein, glycoproteins, and DNA.}{5}
\pairedcasepanel{(f) Model-train animation.}{Match the reference train and camera motion with an editable rig; deliver ${\sim}2.7$\,s at 1080p/30\,fps.}{6}
\caption{\textbf{Construction and verification across another three workflows.}
Opus-5 is on the left and GPT-6-Astra on the right. Both systems use high-to-low
texture baking for the chest and construct two membrane leaflets, with
all binary criteria passing in both tasks. Their checks differ in the
train task: Opus-5 inspects axle/path parameterization, while GPT-6-Astra fits
the camera numerically. Both deliver 80-frame clips at 30\,fps, but GPT-6-Astra
has higher frozen Rubric and Dimension scores. The frames show 0.00,
1.30, and 2.60\,s. }
\label{fig:paired_cases_workflows}
\end{figure}
\clearpage
\endgroup